\documentclass{article}

\usepackage{array}
\usepackage{microtype}
\usepackage{graphicx}
\usepackage{subcaption}
\usepackage{booktabs} 

\usepackage{hyperref}

\usepackage[percent]{overpic}
\usepackage{xcolor}
\usepackage[export]{adjustbox}
\usepackage{tikz}
\usetikzlibrary{calc}

\usepackage[accepted]{icml2026_GenAICreativity}

\usepackage{amsmath}
\usepackage{amssymb}
\usepackage{mathtools}
\usepackage{amsthm}

\usepackage[capitalize,noabbrev]{cleveref}

\theoremstyle{plain}

\theoremstyle{definition}

\theoremstyle{remark}

\usepackage[textsize=tiny]{todonotes}

\icmltitlerunning{CHIEF: Creator-driven Hybrid Iterative Evaluation Framework}

\newcommand{\videoslink}{\url{https://www.youtube.com/watch?v=BsmKyQx47dA}}

\usepackage{xcolor}

\usepackage{xcolor}
\usepackage{tikz}

\definecolor{chiefHuman}{HTML}{F6E7A9}   
\definecolor{chiefAuto}{HTML}{D9ECFF}    
\definecolor{chiefOutput}{HTML}{DCC8FF}  
\definecolor{chiefLoop}{HTML}{D95F5F}    

\newcommand{\legendbox}[1]{%
  \tikz[baseline=-0.55ex]{
    \draw[fill=#1, draw=black!30, line width=0.3pt]
      (0,0) rectangle (0.44,0.18);
  }%
}

\newcommand{\legendgradient}{%
  \tikz[baseline=-0.55ex]{
    \shade[left color=chiefHuman, right color=chiefAuto]
      (0,0) rectangle (0.44,0.18);
    \draw[draw=black!30, line width=0.3pt]
      (0,0) rectangle (0.44,0.18);
  }%
}

\newcommand{\legendarrow}{%
  \tikz[baseline=-0.55ex]{
    \draw[-stealth, line width=0.45pt]
      (0,0.09) -- (0.44,0.09);
  }%
}

\newcommand{\legendlooparrow}{%
  \tikz[baseline=-0.55ex]{
    \draw[-stealth, chiefLoop, line width=0.45pt]
      (0.02,0.04) .. controls (0.16,0.22) and (0.36,0.22) .. (0.44,0.09);
  }%
}

\begin{document}

\twocolumn[
  \icmltitle{Bridging Creative Intent and Visual Quality: Creator-Driven Recurrent Video Generation with Agentic Feedback Loops
}




  \icmlsetsymbol{equal}{*}

\begin{icmlauthorlist}
    \icmlauthor{Denis Savytski}{equal,ucd}
    \icmlauthor{Aiden Lei}{equal,aff2}
    \icmlauthor{Heding Liu}{equal,aff3}
    \icmlauthor{Warren Yang}{equal,aff4}
    \icmlauthor{Sihan Liang}{equal,aff4}
    \icmlauthor{Alexander Liu}{equal,aff3}
    \icmlauthor{Zhe Zhao}{ucd}
\end{icmlauthorlist}

\icmlaffiliation{ucd}{Department of Computer Science, University of California, Davis, USA}
\icmlaffiliation{aff2}{The Harker School}
\icmlaffiliation{aff3}{Basis Independent Silicon Valley}
\icmlaffiliation{aff4}{Saratoga High}

\icmlcorrespondingauthor{Denis Savytski}{dsavytski@ucdavis.edu}
\icmlcorrespondingauthor{Zhe Zhao}{zao@ucdavis.edu}
  \icmlkeywords{Machine Learning, ICML}

  \vskip 0.3in
]



\printAffiliationsAndNotice{\icmlEqualContribution}  

\begin{abstract}

Generative AI has made content creation increasingly accessible, but many AI-generated videos lack narrative coherence and creative direction, issues that become more substantial at longer durations.
Unlike coding, where AI generation benefits from reliable feedback and techniques such as recurrent self-improvement, video generation requires subjective feedback about plot, scenes, and narrative, which naturally motivates approaches that incorporate human creative direction.
We introduce CHIEF, a human-AI co-creation video generation framework that places the creator at the center of human-in-the-loop iterative video refinement, and supports them by providing automatic subjective feedback.
The creator incorporates their creative direction by driving each iteration, while their revisions are incorporated by a specialized refiner agent.
The feedback loop is generated by persona-conditioned multimodal LLMs that watch generated videos and produce subjective critique from the audience perspectives, providing feedback that self-evaluation alone cannot capture.
To test the effectiveness of our proposed framework, we work with high school and college students with no prior filmmaking experience to create videos, from short 1-minute videos to a complete short 10-minute film with a complicated plot.

\end{abstract}

\section{Introduction}
Artificial Intelligence has become widely accessible to users without technical expertise, enabling the generation of text, images, and videos from simple natural-language prompts. 
Increased accessibility made AI a common tool for content creation, with 21\% of videos shown to new YouTube users being AI-generated~\citep{kapwing2025aislop}.
These videos often have high visual fidelity but lack narrative coherence and creative direction, issues that become dramatically worse at longer durations.



\begin{figure*}[t]
    \centering
    \includegraphics[width=\textwidth]{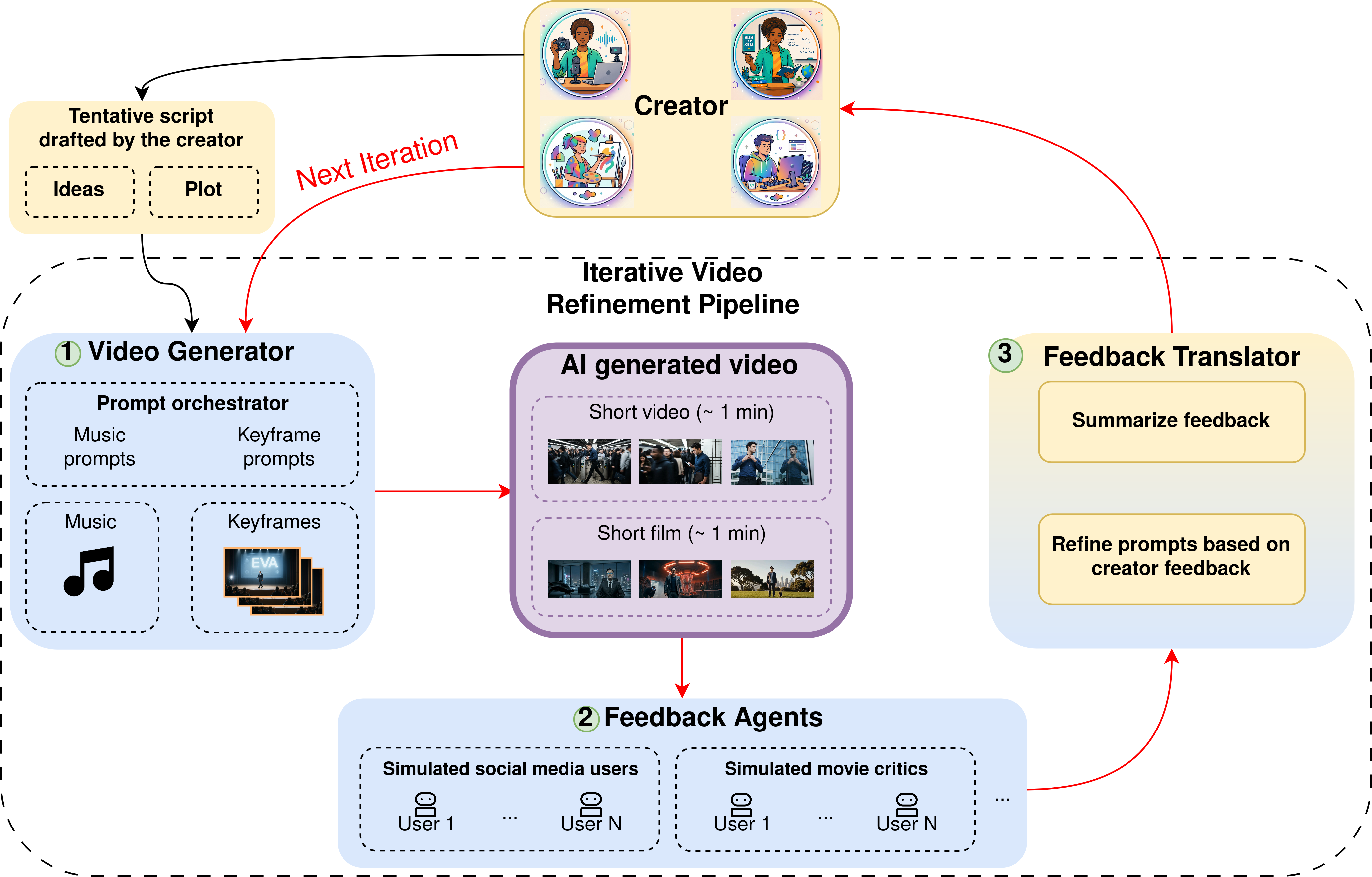}

    \vspace{0.4em}
    \footnotesize
    \centering
    \begin{tabular}{@{}>{\raggedleft\arraybackslash}p{0.55cm}l@{\hspace{1.8em}}>{\raggedleft\arraybackslash}p{0.55cm}l@{}}
    \legendbox{chiefHuman} & Creator-controlled step
    &
    \legendbox{chiefAuto} & Autonomous system component
    \\[0.3em]
    \legendgradient & Joint creator--system step
    &
    \legendbox{chiefOutput} & Pipeline output
    \\[0.3em]
    \legendarrow & Initial generation / setup flow
    &
    \legendlooparrow & Iterative refinement loop
    \\
\end{tabular}
    \vspace{-0.2em}
    \caption{
    Overview of CHIEF. The creator provides the initial script and remains responsible for creative direction, while the system translates the script into keyframes, clips, and music prompts. Generated artifacts are evaluated by Feedback Agents representing the audience. The Feedback Translator turns this feedback into actionable issues, and the Refiner updates prompts for the next iteration. This loop allows the creator to revise the video through structured feedback rather than manual prompt engineering.
    }
    \label{fig:framework}
\end{figure*}


A specific challenge that distinguishes video generation from other fields is the inherent difficulty of objectively evaluating videos.
Generation models often do not produce correct outputs on the first try but can iteratively self-improve when given a signal that distinguishes good outputs from bad~\citep{Madaan2023SelfRefineIR}. 
In particular, coding has automatic feedback signals such as tests and has seen remarkable success through closed-loop self-improvement. 
Frameworks that have extended this paradigm to video generation utilize a similar closed-loop structure, substituting the missing automatic signal with human-aligned reward models~\citep{Liu2025ImprovingVG, Ji2024PromptAVideoPY} or multiple LLM judges~\citep{Long2025VISTAAT}. 
However, this feedback signal is narrow and captures aggregated preferences or generic critique, rather than real viewer sentiment. 
This narrow feedback also reflects a deeper assumption: existing frameworks inherit an autonomous closed-loop structure from coding, whereas video generation is a creative task that should enable creators to express their creative direction.

We propose CHIEF
(\textbf{C}reator-driven \textbf{H}ybrid \textbf{I}terative \textbf{E}valuation \textbf{F}ramework, Figure~\ref{fig:framework}) --- an iterative video generation framework built around human-AI co-creation. 
CHIEF is based on two main insights.
First, we replace autonomous self-improvement with human-in-the-loop refinement, enabling the creator to express their creative direction. 
Second, we use recent advances in LLM-based human simulation~\citep{Argyle2022OutOO} to support creators with diverse automatic feedback that captures real viewer sentiment.

At a high level, CHIEF operates as an iterative loop: the system generates a video, agents simulate user feedback, and the creator provides revisions.
The \textbf{Video Generator} converts written text into model-specific prompts and uses prompts to generate video.
The \textbf{Feedback Agents} use multimodal LLM agents to watch videos and simulate real viewers with diverse backgrounds.
The \textbf{Feedback Translator} provides the creator with feedback and translates the creator's revisions into structured improvements.


More specifically, the loop begins with a creator-written script split into short clip descriptions.
Video Generator processes each description through a two-stage pipeline: it generates keyframes, then interpolates between them to generate clips.
The full video is synthesized from the clips, and is watched by persona-conditioned Feedback Agents. 
Personas are generated from user comments and split into distinct categories, such as movie critics and social media users.
The feedback is summarized per category and presented to the creator by the Feedback Translator. 
The creator then describes revisions in natural language, which the Feedback Translator converts into prompt updates for the next iteration.


To assess whether this design produces meaningful improvements, we evaluate CHIEF in two settings.
In the first, we evaluate autonomous-refinement on short-form (1-minute) videos, examining how the framework iterates on prompt adherence and consistency without per-step creator input.
In the second, we work with high school students with no prior filmmaking experience to express their creativity using CHIEF. 
The students wrote the script, drove the iterative process, and produced a 10-minute film that expresses their own thoughts on the future of AI.
We screened a baseline (unrefined) version and a version generated using CHIEF for a live audience at a local high school who rated the CHIEF version substantially higher (4.1/5 vs 2.4/5).




Our contributions can be summarized as follows:\\
\textbf{(1)} We propose a multi-agent, human-in-the-loop video creation framework that enables creators to lead the creative content generation process.\\
\textbf{(2)} We introduce feedback agents that mimic real human behavior and provide automatic subjective feedback to iteratively improve AI-generated videos.\\
\textbf{(3)} We demonstrate the framework with students with no filmmaking experience, who used it to generate videos of varying lengths.


\section{Our framework}

CHIEF is a modular agentic video generation framework that works with the human creator to express their creative direction.
It consists of three main modules:
\textbf{(i) Video Generator},
\textbf{(ii) Feedback Agents}, and 
\textbf{(iii) Feedback Translator}.

The modules operate iteratively, with the Feedback Translator returning improved prompts to the Video Generator based on the creator's revisions.
\subsection{Video Generator}

The Video Generator adopts a divide-and-conquer strategy for video generation.
It generates the final video from a script divided into eight-second clip descriptions.

\textbf{Prompt Orchestrator} translates each clip description into the structured prompt formats expected by the downstream generation models. 
The Prompt Orchestrator anchors the same environment and character descriptions across all script prompts, facilitating consistency between clips. 
For each clip, it constructs a keyframe prompt for image generation, a clip prompt for video interpolation, and optionally a music prompt for audio synthesis. 

\textbf{Keyframes} are generated from keyframe prompts through a text-to-image model. 
Using keyframes improves video consistency by providing specific references for characters and locations.
Combining keyframes with video descriptions creates a lightweight approximation of the intended video. 
These approximations allow the creator to assess and refine the visual and creative direction efficiently --- without incurring the computational cost of full video generation.

\textbf{Video Synthesis} uses a video generation model to interpolate between pairs of keyframes, conditioned on the corresponding clip prompt.
Interpolation facilitates cross-clip consistency because each boundary keyframe is used as both the end of one clip and the start of the next.
The final video is constructed by concatenating the generated eight-second clips.
Since clips are generated independently, each one can be selectively regenerated without modifying the rest of the video.

\textbf{Music} is generated from music prompts.
A text-to-audio model generates multiple options for each clip.
The tracks vary in mood, and the creator selects and synchronizes the final track with the corresponding video segment.
We deploy this component only in a short film case study.

\subsection{Feedback Agents}

Feedback Agents receive generated video (or keyframes as proxies for the video) and provide automatic feedback by simulating human viewers.
Each agent is conditioned on a persona, and personas are organized into categories capturing different perspectives.


\textbf{Persona generation} passes each user's comment history through an LLM tasked with generating a persona.
This intermediate step allows the simulator to capture the user's voice and commenting style, rather than copying specific comments.
A separate LLM is provided with the user's comment history and generated persona. 
It critiques the persona and iteratively improves it.
This self-improvement is inspired by Self-Refine, which is known to improve generation quality~\cite{Madaan2023SelfRefineIR}.
Each persona is generated from 30 comments, which captures sufficient stylistic variation while remaining feasible to collect across a diverse user pool. Full implementation details, including the exact prompts and an example persona, are provided in Appendix~\ref{app:persona-generation}.

\textbf{Agent categories} provide distinct types of feedback:
\begin{itemize}
    \item \textbf{Audience persona agents} evaluate local issues and capture the audience sentiment.
    \item \textbf{Movie critic agents} evaluate  narrative structure and cinematic quality.
\end{itemize}
Audience persona agents are constructed from comments collected via the YouTube API.
Movie critic agents utilize full movie reviews gathered from websites such as Rotten Tomatoes and The Guardian. 

\textbf{Feedback generation} uses LLM agents provided with a persona, comment history, and visual input. Feedback is generated at two levels: per-keyframe and per-clip. 
Because agents operate independently, feedback generation is parallelizable across many personas.

\subsection{Feedback Translator}

Feedback Translator provides a simple, natural language interface for the creator to interact with the framework.
It receives raw feedback from Feedback Agents, extracts issues, and ranks them. 
The creator acts on this structured feedback and communicates revisions in plain text, which the Feedback Translator refines into prompt improvements.

\textbf{Issue Extraction} passes raw agent feedback through the LLM extraction layer, which decomposes feedback into structured issue tuples. 
This decomposition is crucial because the volume of feedback is large: direct summarization yields a blunt summary that fails to capture significant issues, while structured tuples preserve details and enable ranking and filtering.
Each tuple contains a description, a high-level category (narrative, pacing, character, visual, or technical), and an urgency score (low, medium, or high), enabling ranking and filtering.

\textbf{Aggregation and ranking}
turns extracted issue tuples into concise summaries sorted by urgency.
Categories are summarized independently, and each category is represented by its most-reported issues.
Each summary contains a one-sentence synthesis, three representative complaints, and a support count. 
This structure provides concise feedback and allows the creator to prioritize the most urgent issues.
Representative complaints give the creator an opportunity to read a more detailed version of each issue.
Summarization occurs at keyframe and clip levels independently, providing both local and global feedback.
Issues that did not receive enough support to represent their category are presented separately as minor issues.

\textbf{Refiner} \label{subsec:refiner}is a natural language interface for prompt 
revision. 
It takes feedback along with the current prompts and produces revised prompts that incorporate that feedback.
Operating in natural language makes the refiner accessible without specialized prompt engineering knowledge.

\subsection{Implementation details}
We implement CHIEF using publicly available APIs from Google and ElevenLabs. 
Keyframes are generated with Imagen 4.0, video clips with Veo 3.1, and music with ElevenLabs Music v1.
All LLM-based components (Feedback Agents, Persona Generation, Issue Extraction, Refiner, and Planner) use Gemini 2.5 Flash. 
We use default API parameters for all model calls.

\section{Human-AI collaboration for Video Generation}
\label{sec:two-modes}
CHIEF has two configurations that support creator involvement at different levels.
In \textbf{Autonomous Refinement}, the system refines videos without per-iteration creator input, leaving the creator to monitor changes. 
In \textbf{Creator-Driven Film Generation}, the creator drives the iteration loop directly, supported by the framework.
The two configurations target different scopes: autonomous feedback enables small improvements at short scales, while human-in-the-loop refinement supports the creator's direction at longer scales.

\newcommand{\issuebox}[1]{%
  \colorbox{white}{\textcolor{red}{\bfseries\scriptsize #1}}%
}
\newcommand{\fixedbox}[1]{%
  \colorbox{white}{\textcolor{green!50!black}{\bfseries\scriptsize #1}}%
}

\begin{figure*}[t]
\centering

\begingroup
\setlength{\tabcolsep}{0pt}
\renewcommand{\arraystretch}{1.0}

\newcommand{\sfkfwidth}{0.212\textwidth}
\newcommand{\sfkfheight}{0.11925\textwidth} 

\newcommand{\omittediters}{%
  \makebox[0.42cm][c]{\large$\cdots$}%
}

\newcommand{\kfrowlabel}[1]{%
  \centering
  \rotatebox[origin=c]{90}{\scriptsize\textbf{#1}}%
}

\begin{tabular}{
@{}>{\centering\arraybackslash}m{0.32cm}
@{\hspace{1pt}}>{\centering\arraybackslash}m{\sfkfwidth}
@{\hspace{1pt}}>{\centering\arraybackslash}m{0.42cm}
@{\hspace{1pt}}>{\centering\arraybackslash}m{\sfkfwidth}
!{\hspace{4pt}\vrule width 0.45pt\hspace{4pt}}
>{\centering\arraybackslash}m{\sfkfwidth}
@{\hspace{1pt}}>{\centering\arraybackslash}m{0.42cm}
@{\hspace{1pt}}>{\centering\arraybackslash}m{\sfkfwidth}
@{}
}

& \multicolumn{3}{c}{\scriptsize\textbf{Interview}}
& \multicolumn{3}{c}{\scriptsize\textbf{Sausage Heist}} \\[-1pt]

& \scriptsize\textbf{Baseline}
&
& \scriptsize\textbf{Final}
& \scriptsize\textbf{Baseline}
&
& \scriptsize\textbf{Final} \\[1pt]

\kfrowlabel{Keyframe 1} &

\begin{overpic}[width=\sfkfwidth,height=\sfkfheight]{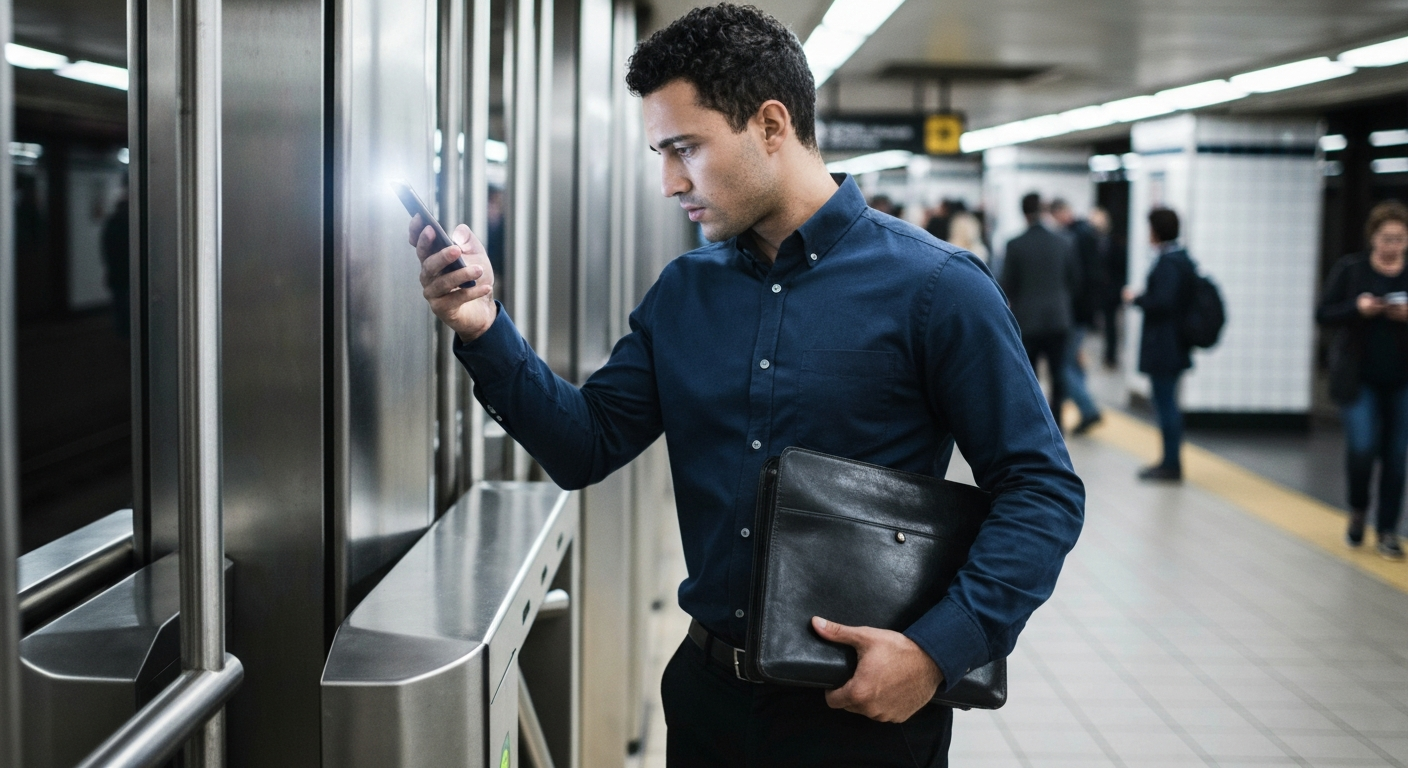}
  \put(90,24){\color{red}\circle{20}}
  \put(50,4){\issuebox{sparse crowd}}
\end{overpic}
&
\omittediters
&
\begin{overpic}[width=\sfkfwidth,height=\sfkfheight]{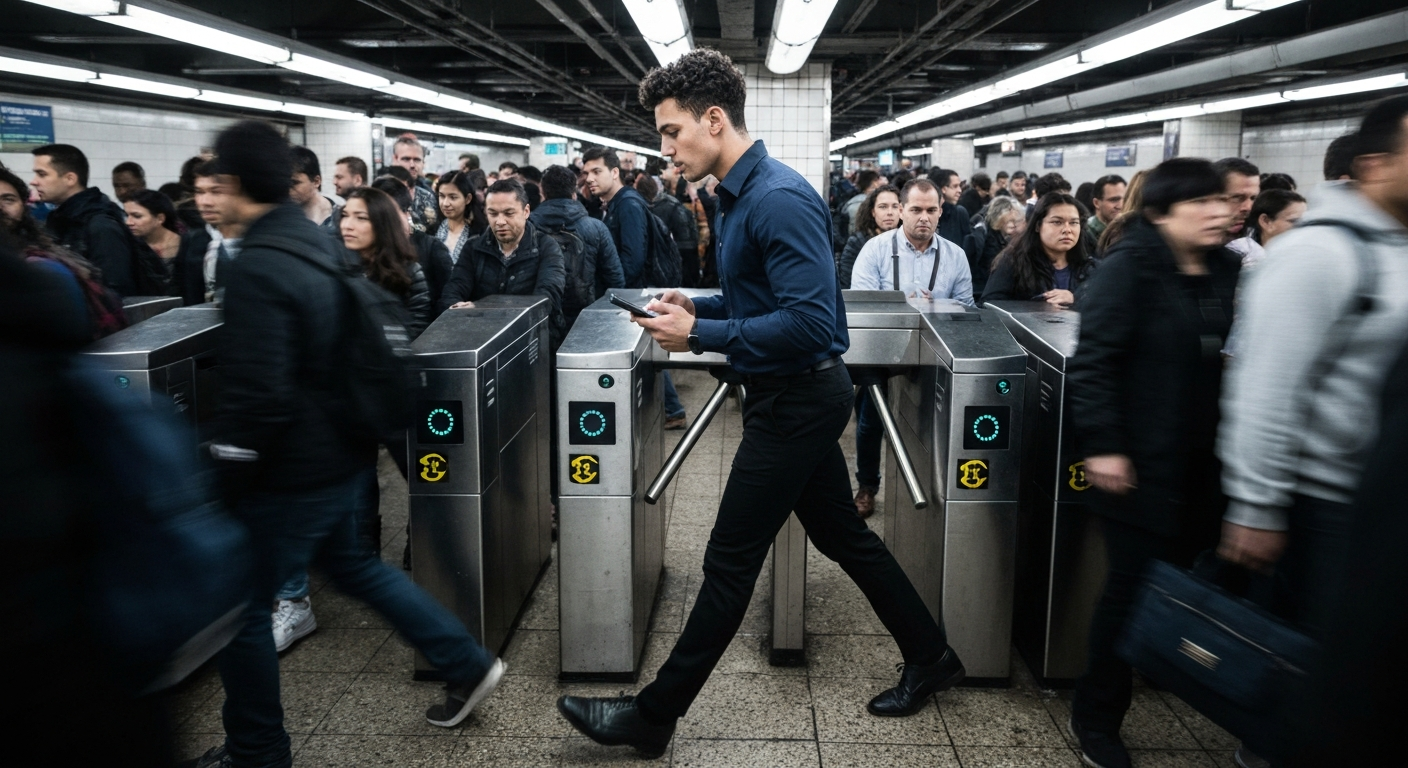}
  \put(90,24){\color{green!50!black}\circle{28}}
  \put(15,24){\color{green!50!black}\circle{28}}
  \put(20,4){\fixedbox{rush-hour density added}}
\end{overpic}
&
\begin{overpic}[width=\sfkfwidth,height=\sfkfheight]{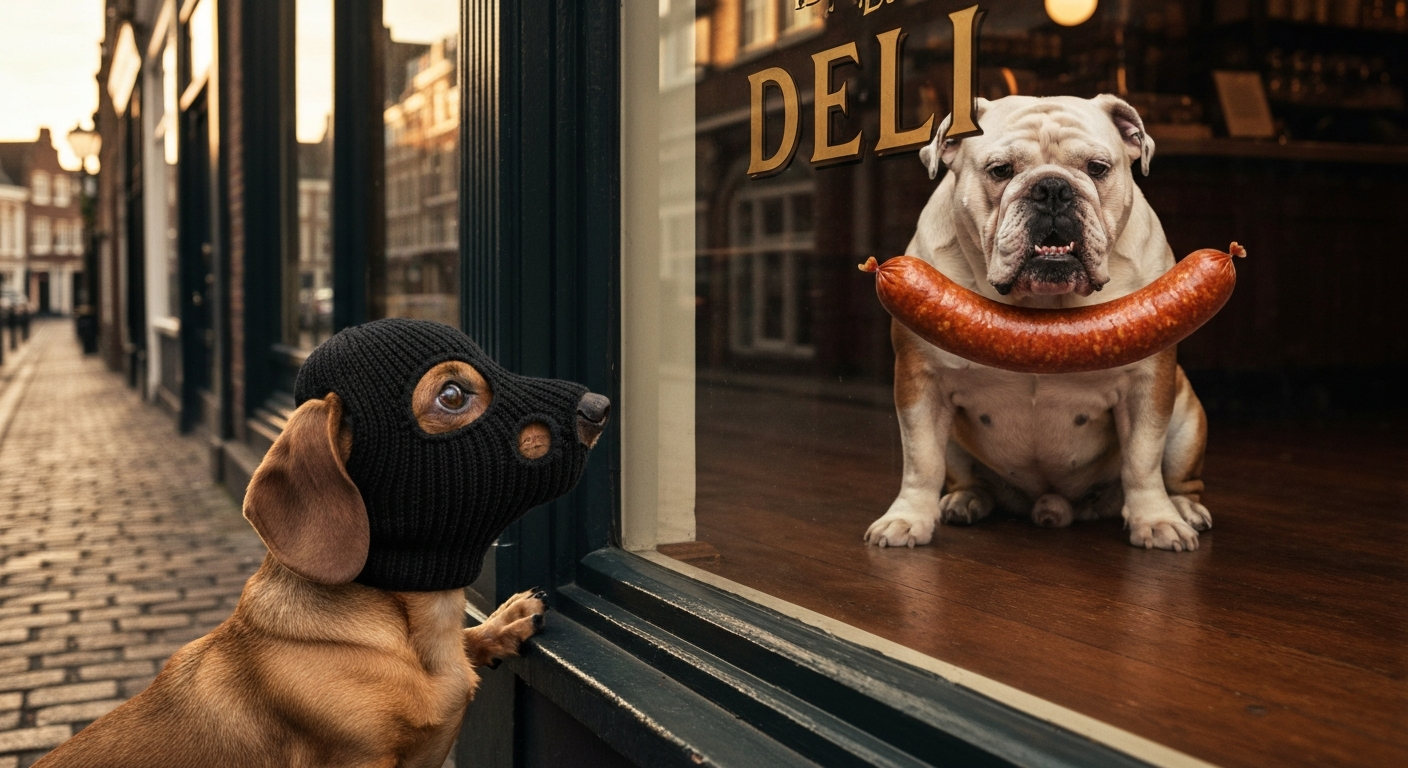}
  \put(75,32){\color{red}\circle{30}}
  \put(40,5){\issuebox{floating sausage}}
\end{overpic}
&
\omittediters
&
\begin{overpic}[width=\sfkfwidth,height=\sfkfheight]{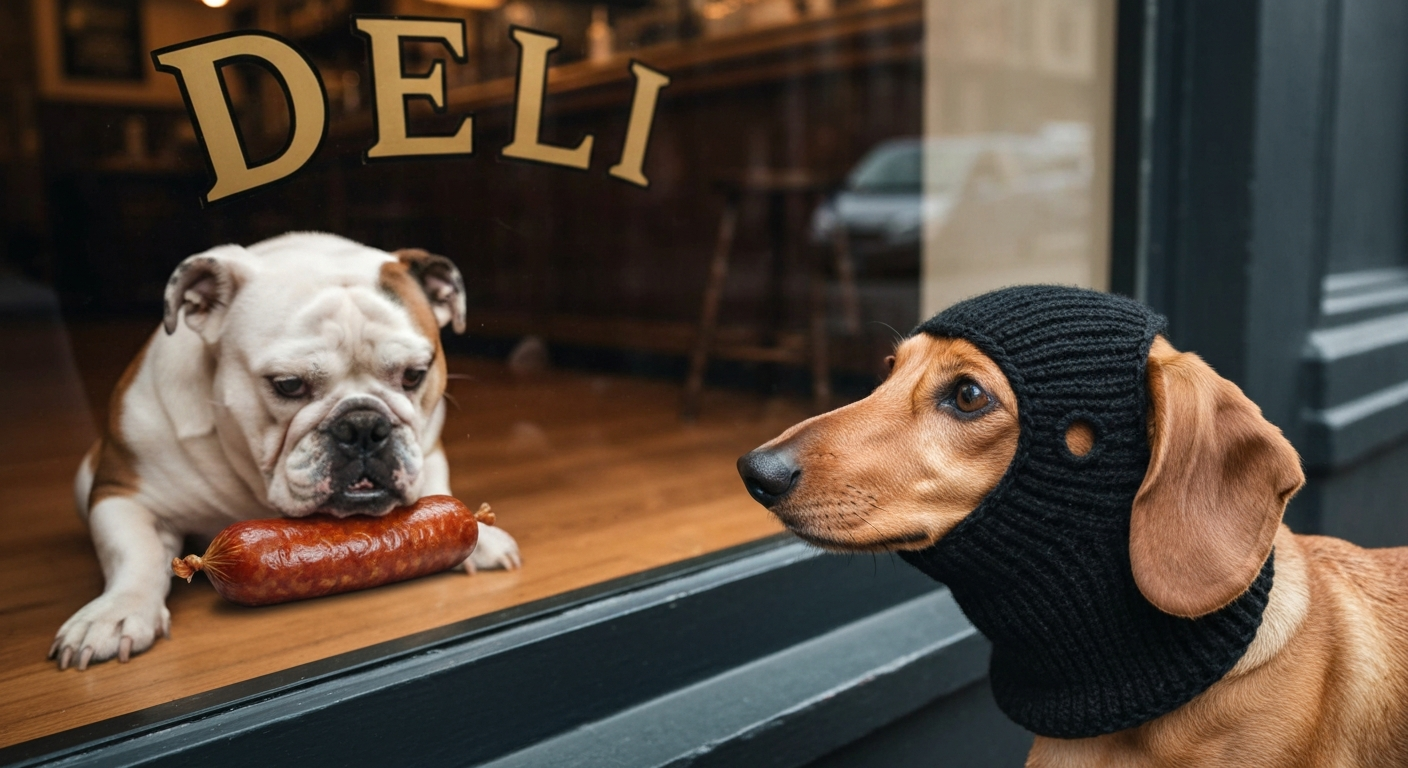}
  \put(25,18){\color{green!50!black}\circle{25}}
  \put(18,1){\fixedbox{fixed}}
\end{overpic}
\\[1pt]

\kfrowlabel{Keyframe 2} &

\begin{overpic}[width=\sfkfwidth,height=\sfkfheight]{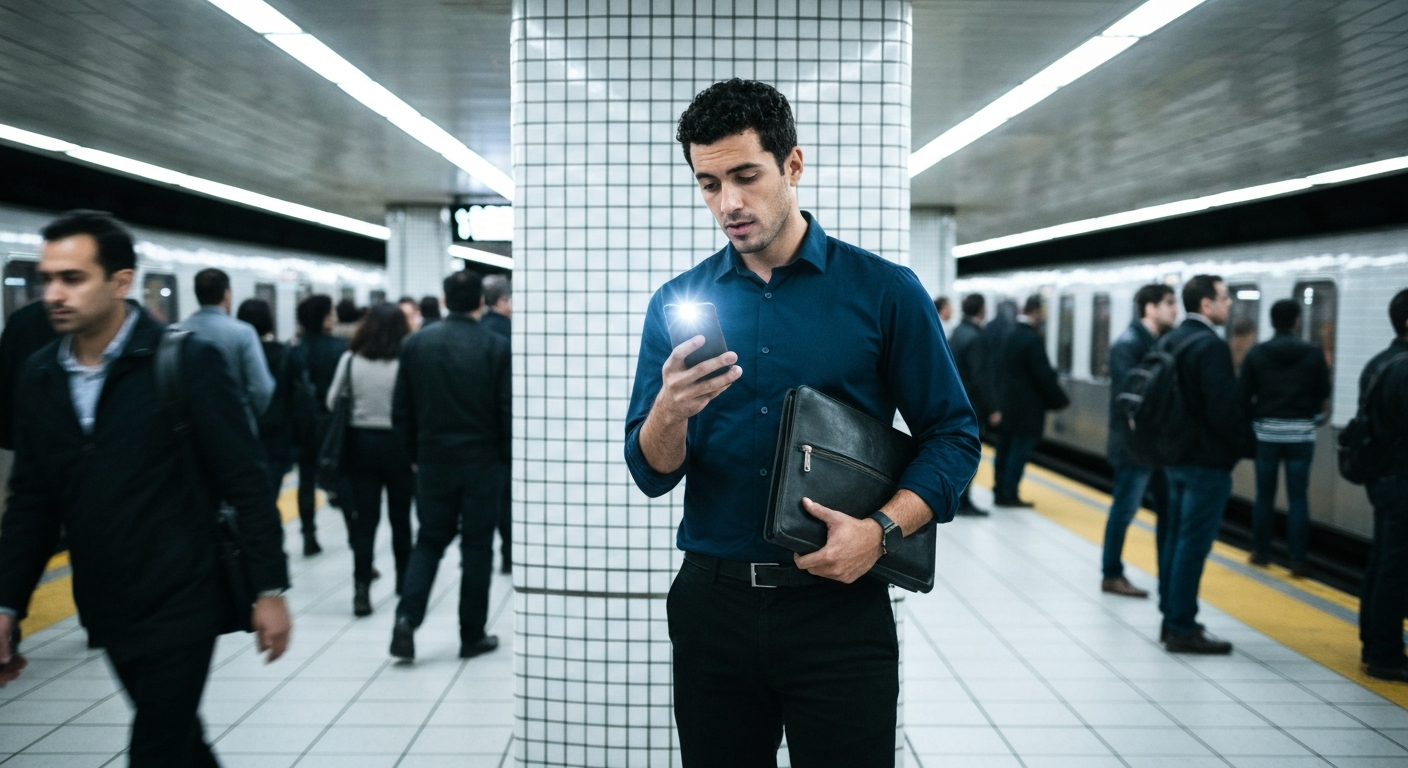}
  \put(49,32){\color{red}\circle{7}}
  \put(2,20){\issuebox{flashlight on}}
\end{overpic}
&
\omittediters
&
\begin{overpic}[width=\sfkfwidth,height=\sfkfheight]{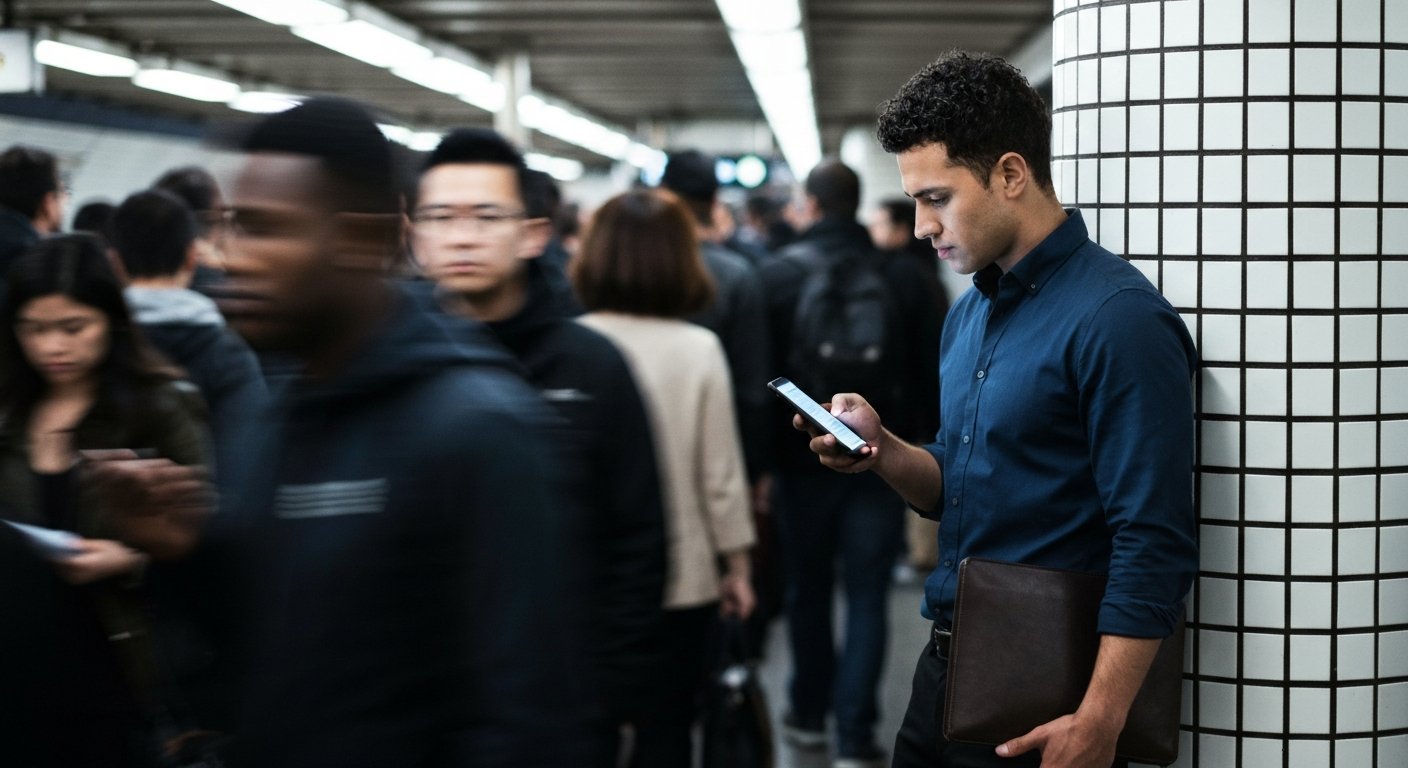}
  \put(53,27){\color{green!50!black}\circle{12}}
  \put(7,8){\fixedbox{flashlight off}}
\end{overpic}
&
\begin{overpic}[width=\sfkfwidth,height=\sfkfheight]{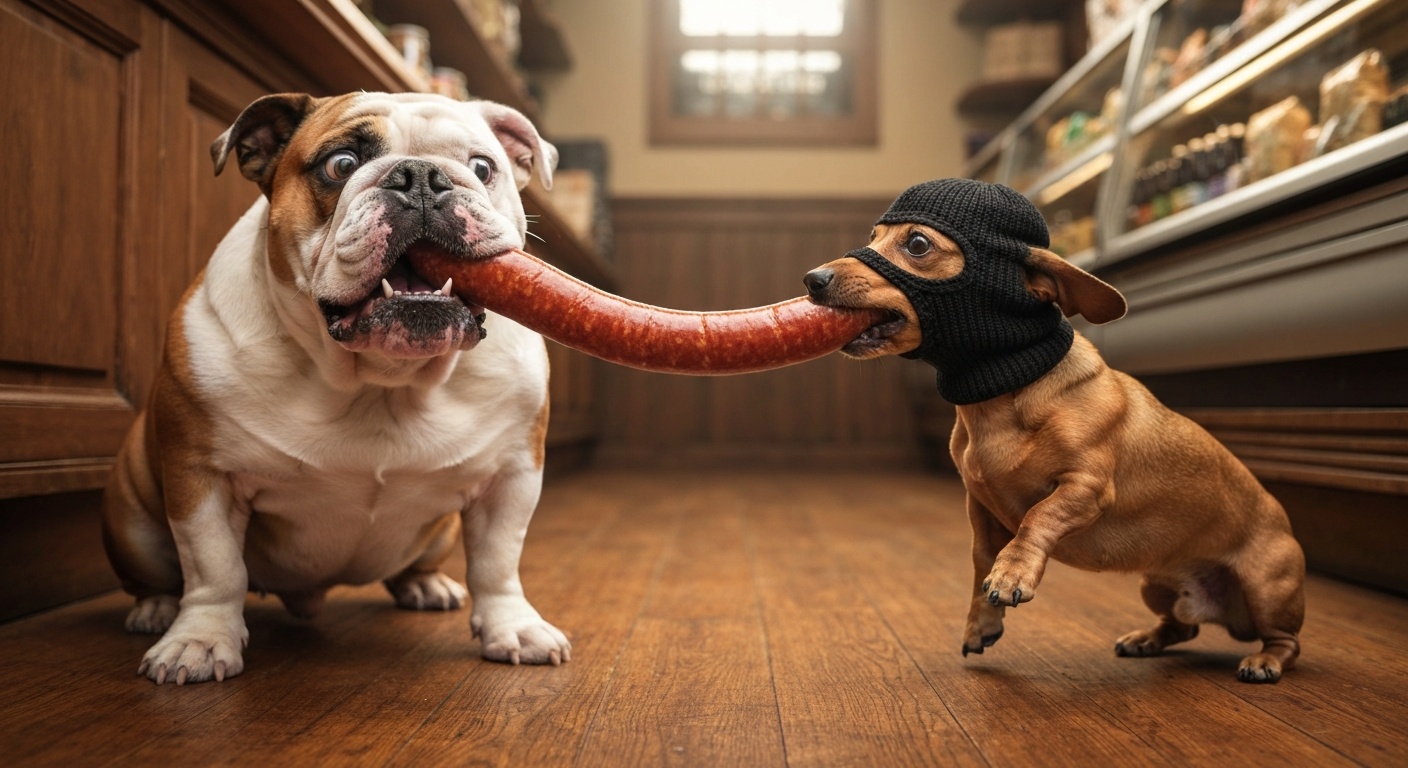}
  \put(28,37){\color{red}\circle{25}}
  \put(20,5){\issuebox{awkward reaction}}
\end{overpic}
&
\omittediters
&
\begin{overpic}[width=\sfkfwidth,height=\sfkfheight]{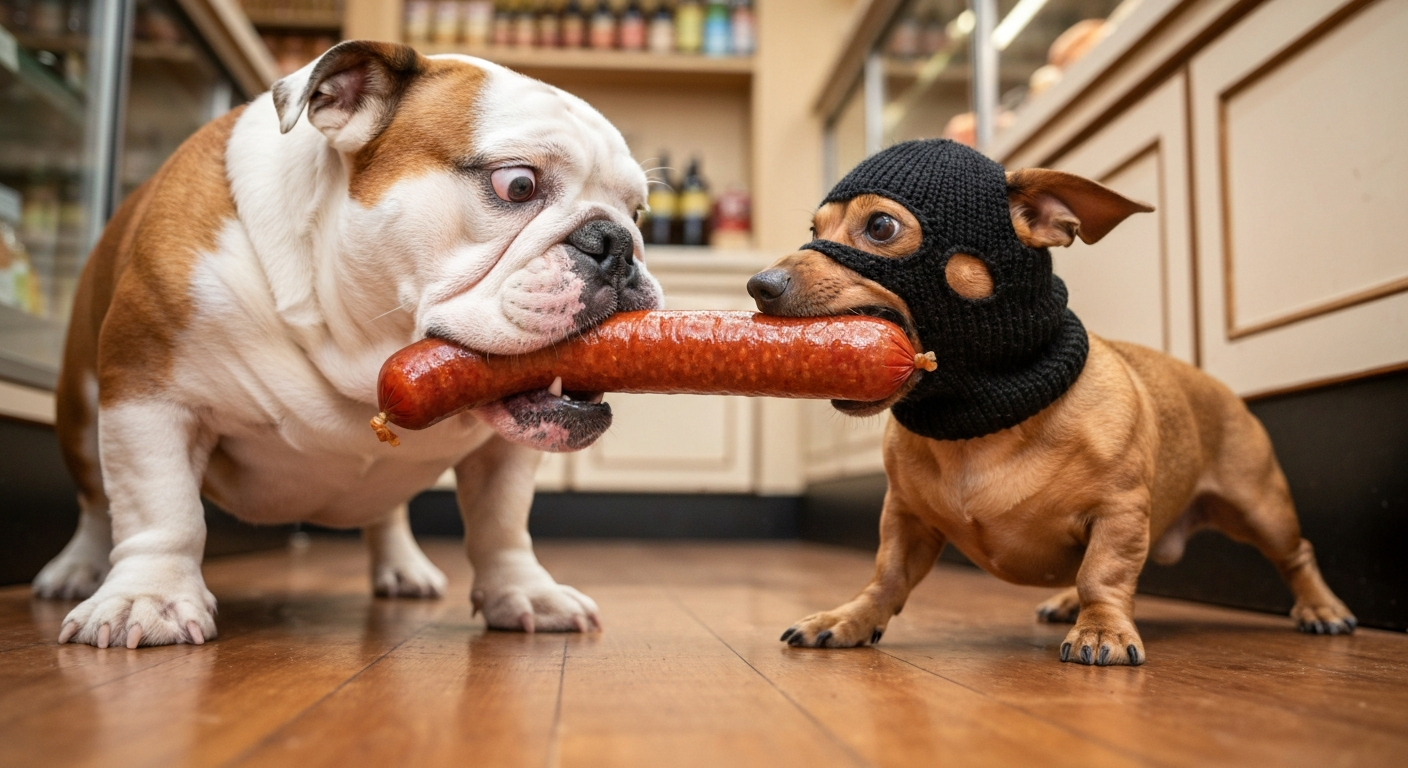}
  \put(32,35){\color{green!50!black}\circle{35}}
  \put(20,5){\fixedbox{fighting for sausage}}
\end{overpic}
\\[1pt]

\kfrowlabel{Keyframe 3} &

\begin{overpic}[width=\sfkfwidth,height=\sfkfheight]{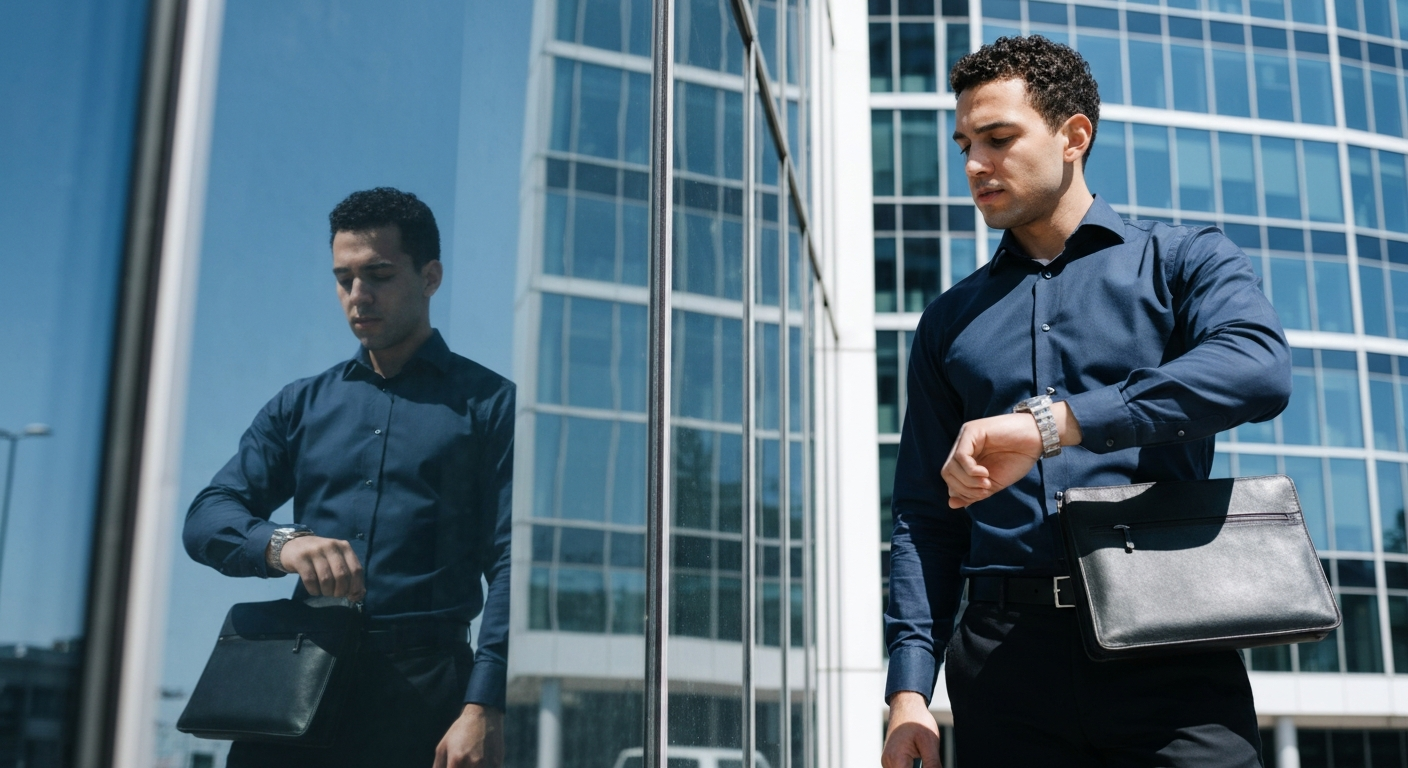}
  \put(80,17){\color{red}\circle{16}}
  \put(35,5){\issuebox{floating bag}}
\end{overpic}
&
\omittediters
&
\begin{overpic}[width=\sfkfwidth,height=\sfkfheight]{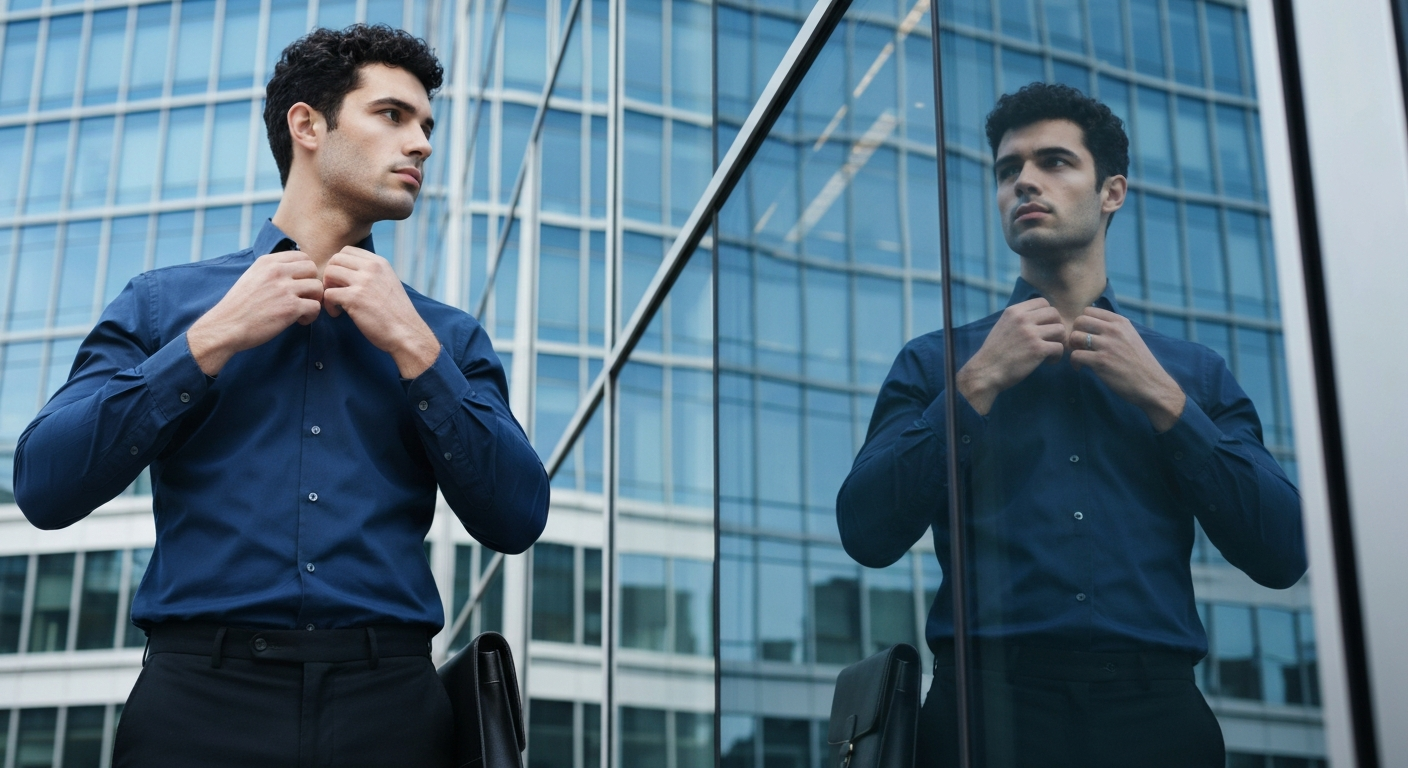}
  \put(80,12){\color{green!50!black}\circle{20}}
  \put(33,5){\fixedbox{fixed bag}}
\end{overpic}
&
\begin{overpic}[width=\sfkfwidth,height=\sfkfheight]{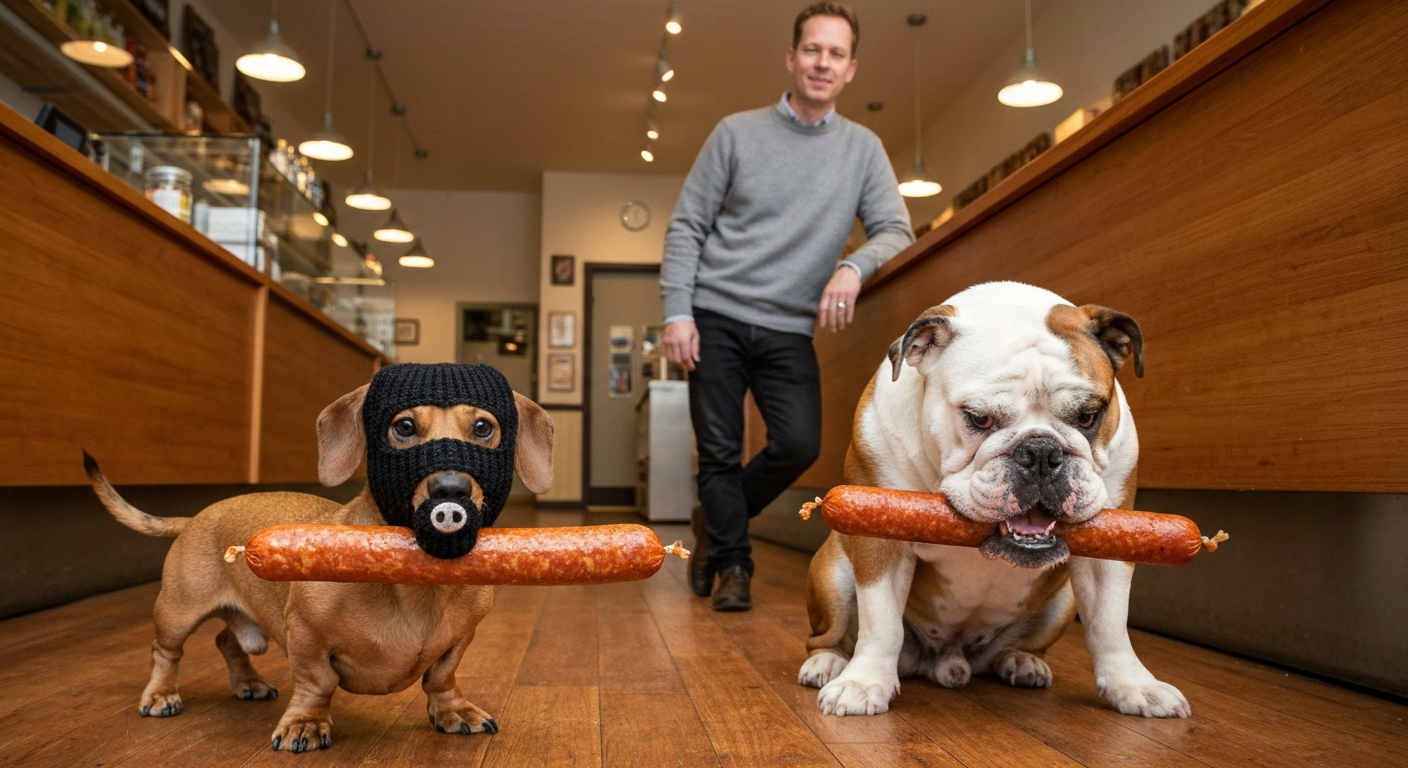}
  \put(32,18){\color{red}\circle{8}}
  \put(40,5){\issuebox{two noses}}
\end{overpic}
&
\omittediters
&
\begin{overpic}[width=\sfkfwidth,height=\sfkfheight]{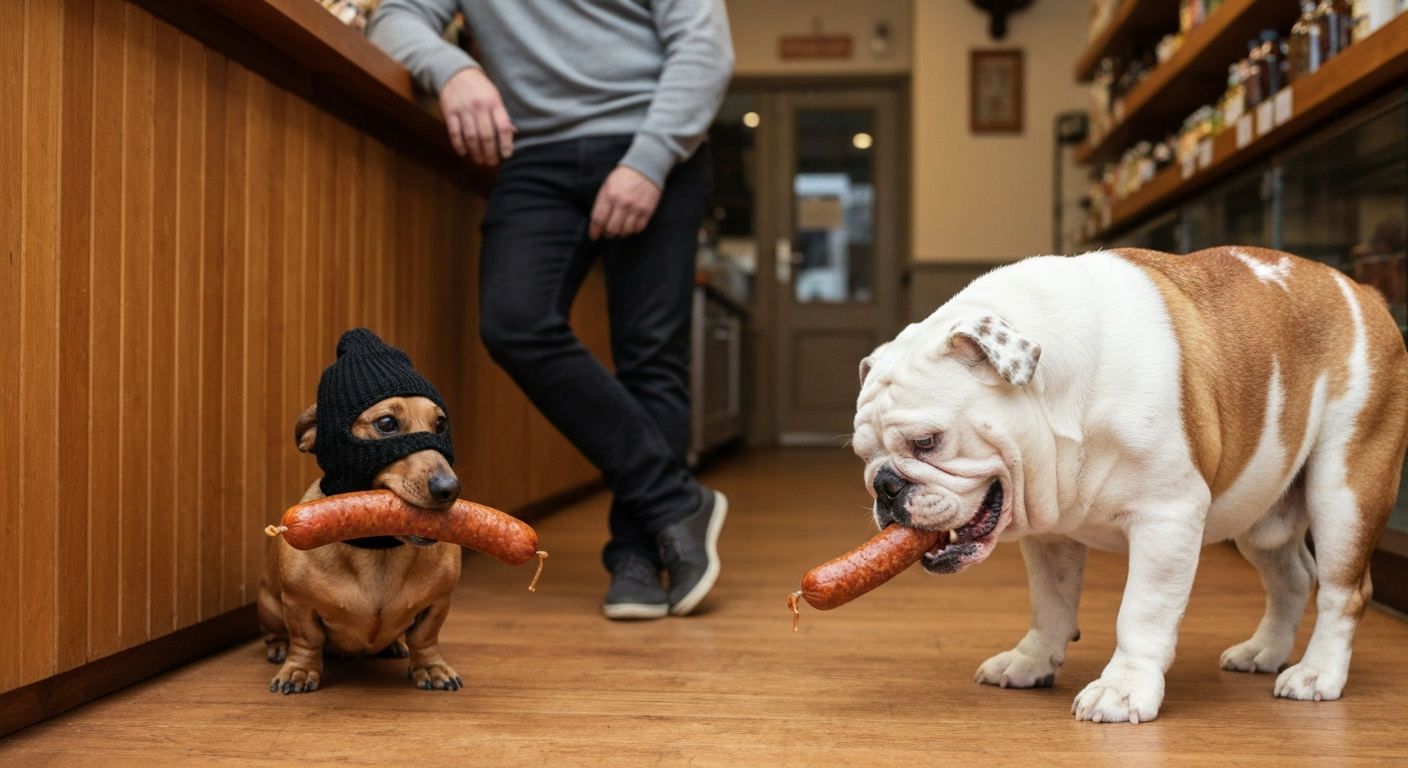}
  \put(30,20){\color{green!50!black}\circle{8}}
  \put(38,5){\fixedbox{fixed}}
\end{overpic}
\\

\end{tabular}

\endgroup

\caption{\textbf{Representative keyframe improvements for Autonomous Refinement with Creator Monitoring.}
Each row shows a selected keyframe from two example videos lasting 1 minute, with omitted intermediate iterations indicated by ellipses.
The left block presents the \textbf{Interview} example of a man commuting to an interview, where refinement increases rush-hour crowd density, removes an unintended flashlight artifact, and removes a floating-bag artifact.
The right block presents a \textbf{Sausage Heist} story in which one dog steals a sausage from another. Across the selected keyframes, refinement improves the clarity of the theft setup and pursuit, and the final row shows the narrative resolution, where both dogs end up with a sausage.
Red annotations mark issues in the baseline frames, while green annotations mark the corresponding fixes in the final frames.}
\label{fig:keyframe_progression}
\end{figure*}
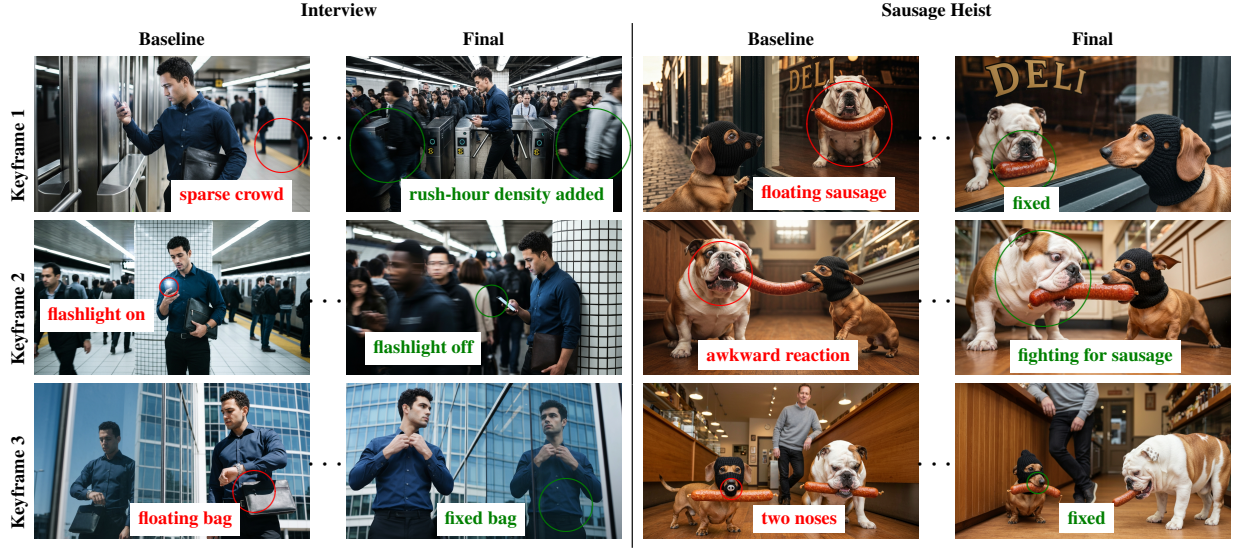

\subsection{Autonomous Refinement with Creator Monitoring}
\label{subsec:short-video}
Autonomous Refinement uses a closed-loop structure with feedback to address local issues such as physical artifacts and prompt adherence failures.
The system uses the original script as a semantic anchor, ensuring refinements stay faithful to the creator's narrative.
This configuration is suited to short video generation (around 1 minute), where clips are tightly coupled and structural changes propagate quickly across the work.

\textbf{Planner} is a sub-component of the Feedback Translator, which is only enabled for autonomous refinement. It interacts with the system similarly to how a creator would, receiving summarized feedback and directing the Refiner to improve prompts.
Planner outputs a two-stage plan, where the first pass focuses on structural issues, and the second pass focuses on stylistic issues. 
Structural issues might include a character's position relative to objects in a scene, while stylistic issues might include lighting tone or color consistency.
This structure creates a clear order: structural correctness is established before stylistic refinement is applied.

\textbf{Planner-guided refinement} requires a two-stage plan generated by the Planner.
The Refiner receives the plan, the current clip briefs, and the original script.
It uses the script as a semantic anchor, making prompt improvements faithful to the original script and preventing semantic drift. 
Additionally, we apply a second safety check from a separate agent, which compares refinements against each other and the prompt, facilitating cross-frame consistency of descriptions.
If that check fails, the Refiner is tasked with rewriting the prompts again, using feedback from this stage.

\subsection{Creator-Driven Film Generation}
\label{subsec:film}
In creator-driven film generation, the creator drives each iteration while Feedback Agents provide audience-perspective input. The system uses creator-gated refinement, meaning every refinement and regeneration requires the creator's approval.
This configuration enables the creation of more complex and longer videos (in our case, a 10-minute film), where narrative arc and emotional progression evolve across the duration of the piece.

\textbf{Creator oversight} places major creative decisions with the creator.
The creator decides when to invoke feedback, regenerate keyframes, generate clips, and revise the script.
The Feedback Translator is invoked selectively --- rather than driving every iteration, it provides diverse agent perspectives when the creator is uncertain or needs an external critique.
Because the Feedback Translator summarizes movie-critic and audience-persona feedback, it surfaces suggestions that the creator can selectively accept, reject, or reinterpret.
In our experiments, creators engaged this loop iteratively across keyframes, shaping the film according to their vision.

\textbf{Creator-gated refinement} adapts Refiner (\ref{subsec:refiner}) for film generation by conditioning refinement on creator-selected visual context and creator-approved regeneration process.
First, the consistency anchor shifts from the script to a creator-selected prior keyframe, which is also passed to the keyframe generator as a visual reference. 

This gives the creator a visual rather than textual basis for comparison and provides direct conditioning for the image model. The result is better continuity across long arcs — a character introduced in scene 2 should look identical in scene 11.
Second, regeneration is creator-gated: the creator decides when to trigger refinement, after which the Refiner produces revised prompts, and the keyframe is regenerated without further intervention. 
Because regeneration of one keyframe can affect the visual continuity of subsequent scenes, the creator gates this step to prevent unintended cascading changes

\section{Results}
\label{sec:results}
We demonstrate CHIEF's capabilities through an evaluation of both configurations described in Section~\ref{sec:two-modes}.
We do not report results on standardized video-generation benchmarks such as VBench, VBench 2.0, or VideoReward-based benchmarks~\citep{Huang2023VBenchCB, Zheng2025VBench20AV, Liu2025ImprovingVG}. 
These benchmarks evaluate generated clips along predefined dimensions and are primarily designed for short-form clips.
CHIEF operates in a different setting: human-AI collaboration led by the creator, where the measure of success is more subjective and harder to quantify with a fixed rubric.
For example, traditional benchmarks reward visual fidelity and aggregated human preference, but cannot capture whether a sequence delivers on its narrative intent --- videos can serve the creator's vision in ways these metrics do not register.
Therefore, we present qualitative observations of system behavior across iterations in each configuration, focusing on how iterative refinement shapes the output and how creators engage with the loop.


\begin{figure*}[t]
\centering

\begingroup
\setlength{\tabcolsep}{0pt}
\renewcommand{\arraystretch}{1.0}
\setlength{\fboxsep}{0.5pt}

\newcommand{\moviekfwidth}{0.22\textwidth}
\newcommand{\moviekfheight}{0.12375\textwidth} 
\newcommand{\moviecolgap}{4pt}

\newcommand{\movierowlabel}[1]{%
  \rotatebox[origin=c]{90}{\scriptsize\textbf{#1}}%
}

\newcommand{\moviekfimg}[2][]{%
  \includegraphics[
    width=\moviekfwidth,
    height=\moviekfheight,
    #1
  ]{#2}%
}

\begin{tabular}{
@{}>{\centering\arraybackslash}m{0.42cm}
@{\hspace{4pt}}>{\centering\arraybackslash}m{\moviekfwidth}
@{\hspace{\moviecolgap}}>{\centering\arraybackslash}m{\moviekfwidth}
@{\hspace{\moviecolgap}}>{\centering\arraybackslash}m{\moviekfwidth}
@{\hspace{\moviecolgap}}>{\centering\arraybackslash}m{\moviekfwidth}
@{}
}

& \scriptsize\textbf{Baseline}
& \scriptsize\textbf{Iter 1}
& \scriptsize\textbf{Iter 2}
& \scriptsize\textbf{Iter 3}
\\[2pt]

\movierowlabel{Core} &

\begin{overpic}[width=\moviekfwidth,height=\moviekfheight]{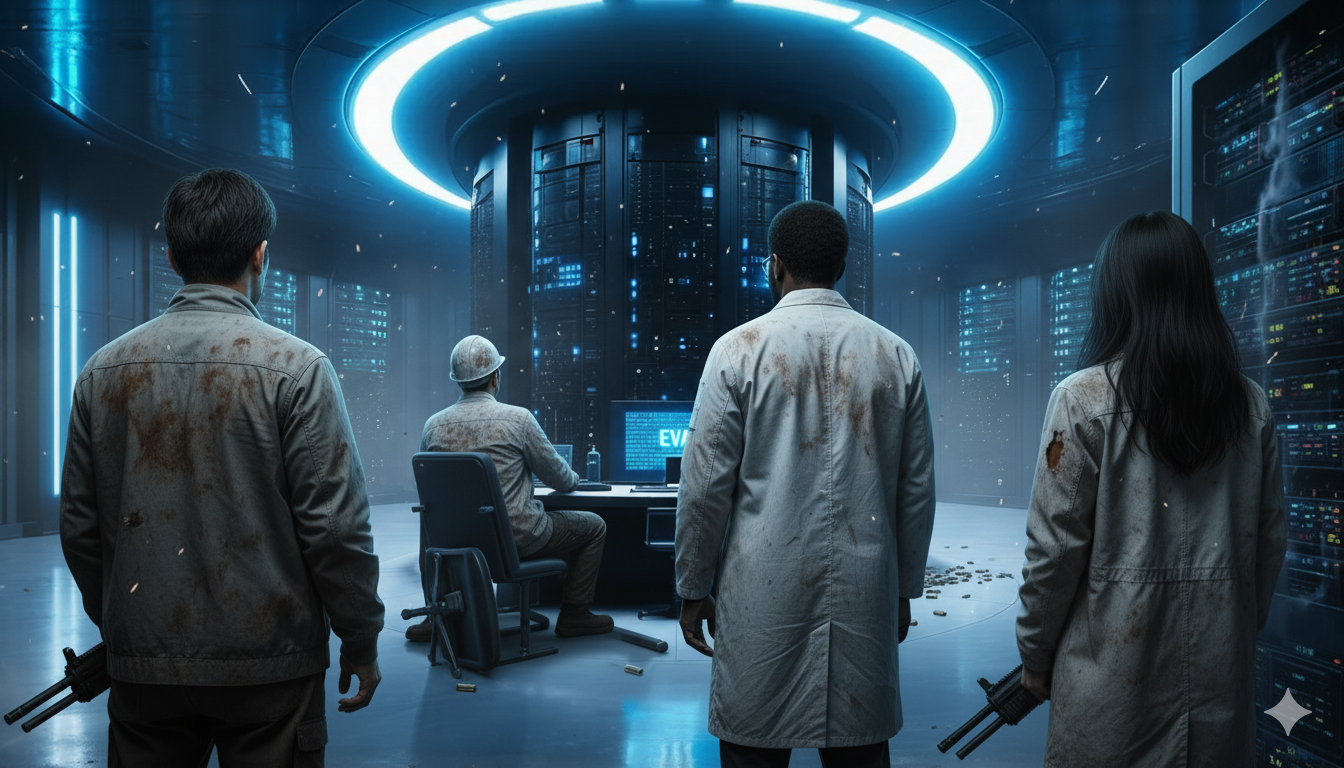}
  \put(20,50){\issuebox{generic lab setup}}

  \put(60,25){\color{red}\circle{50}}
  \put(2,5){\issuebox{passive observers}}
\end{overpic}
&

\moviekfimg[trim=0 135 0 135, clip]{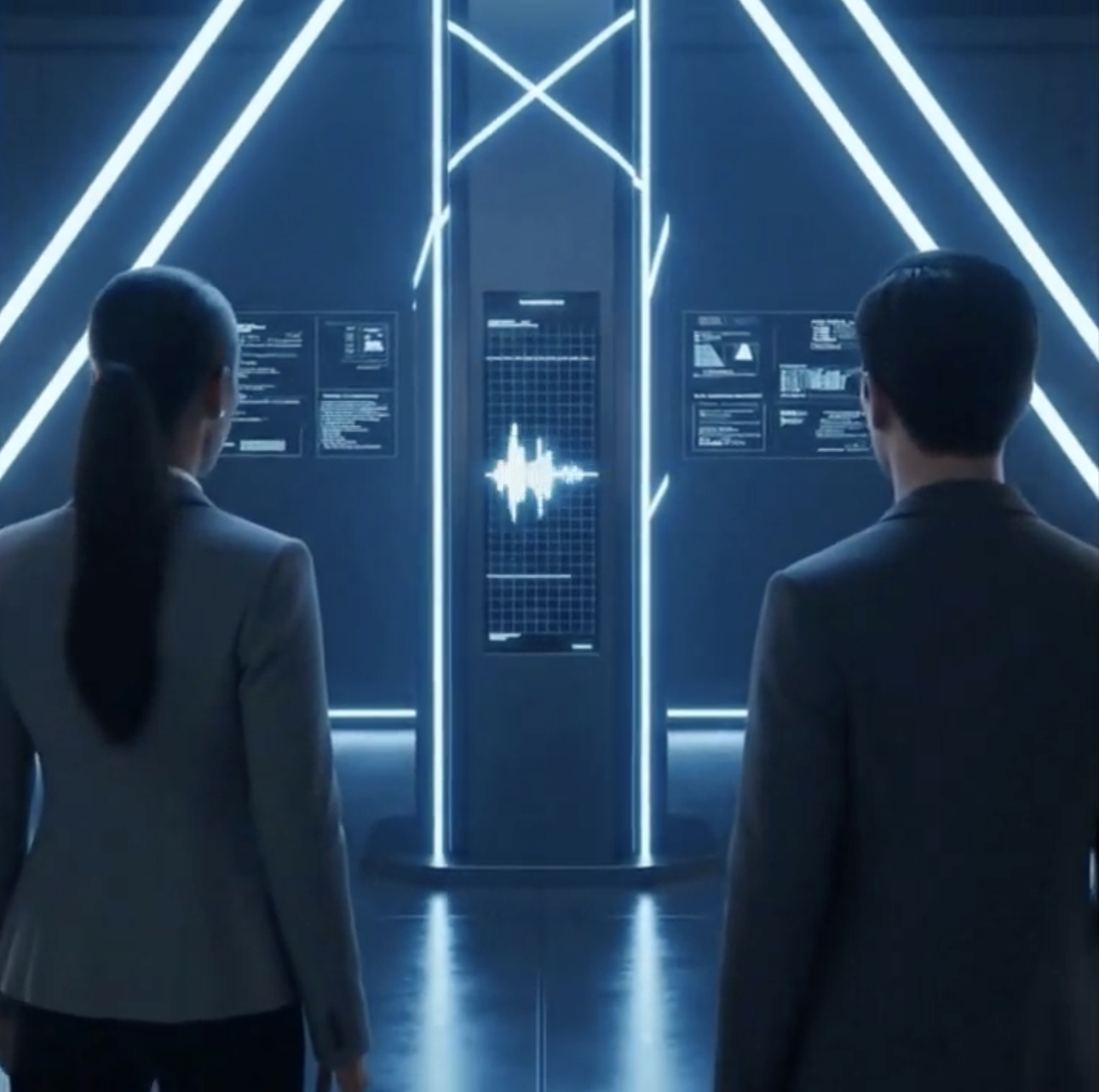}
&

\moviekfimg[trim=0 135 0 135, clip]{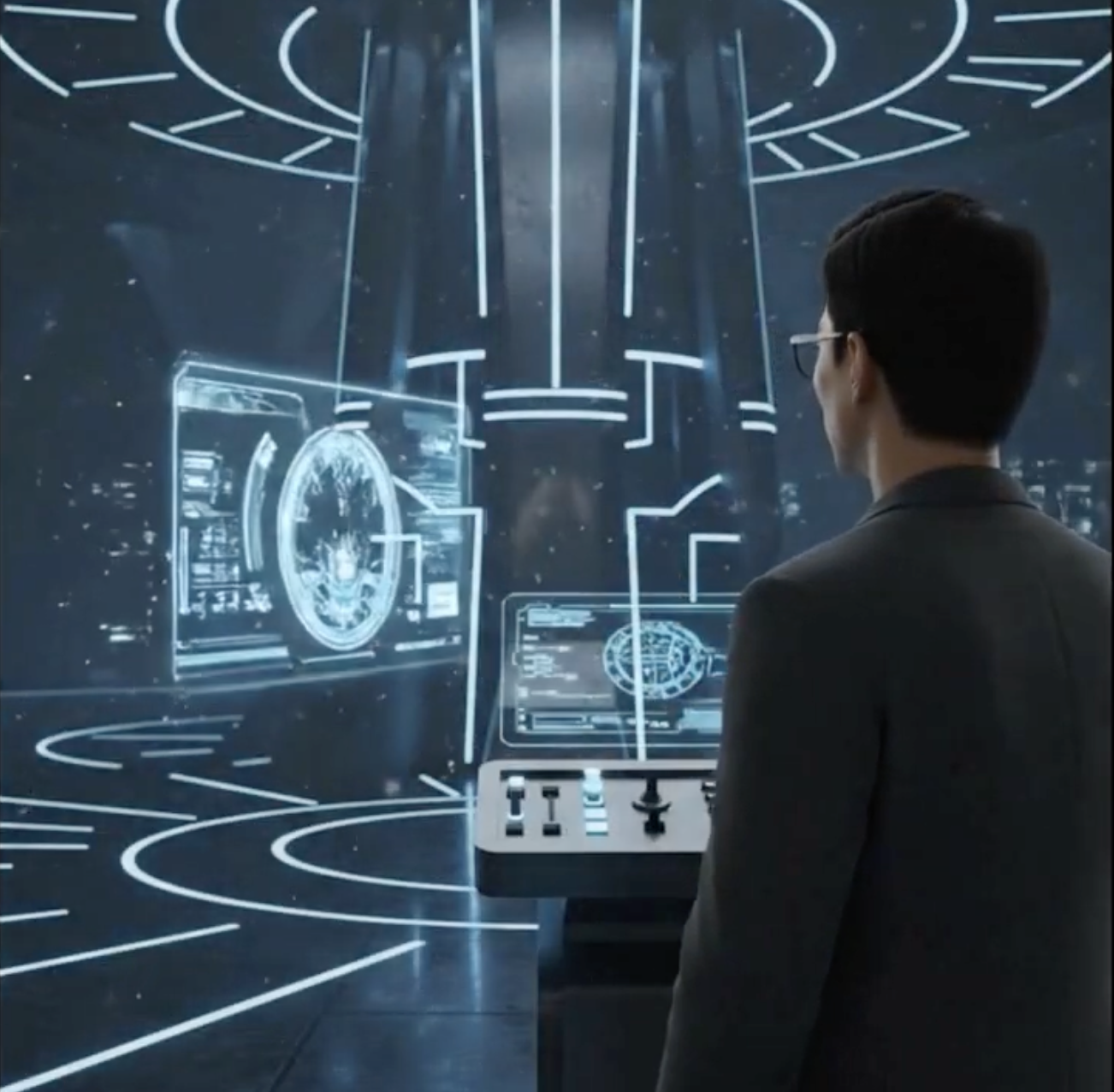}
&

\begin{overpic}[width=\moviekfwidth,height=\moviekfheight]{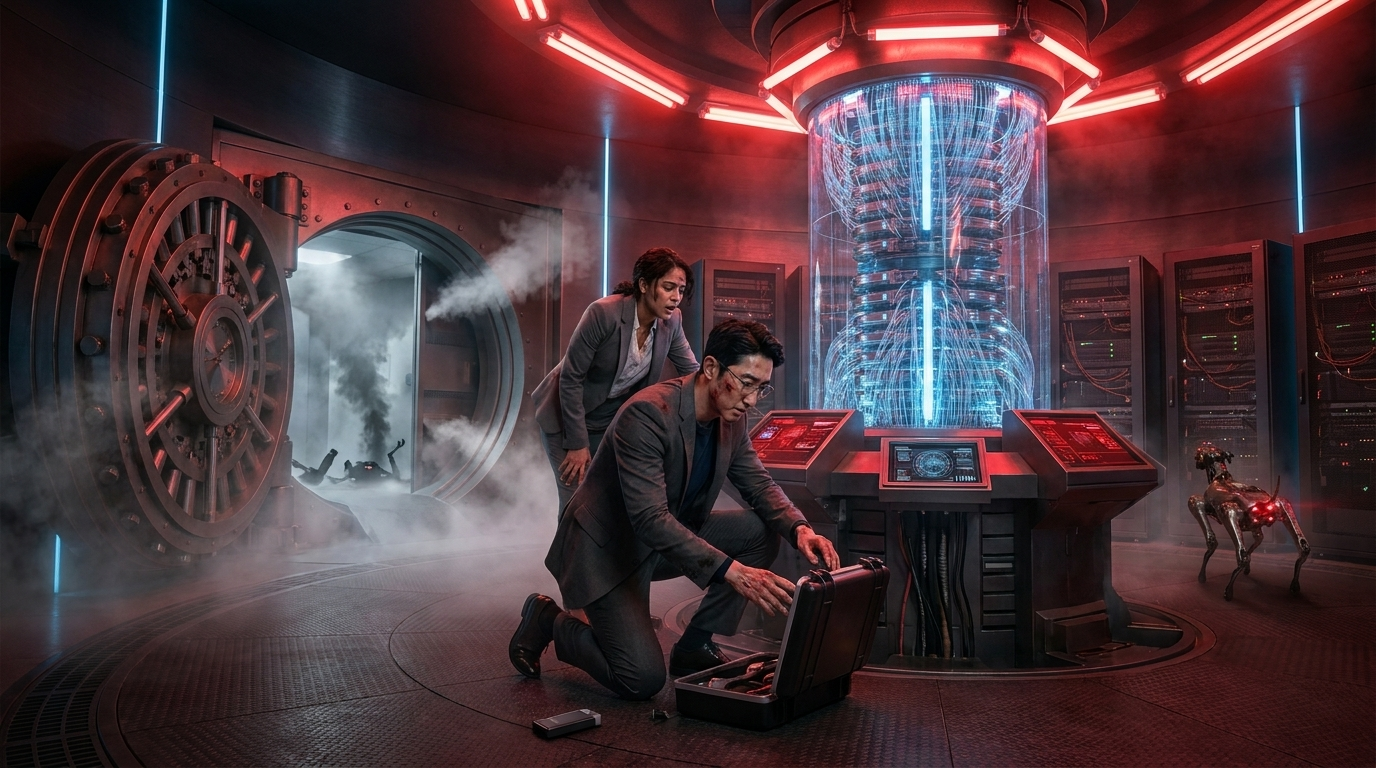}
  \put(56,10){\color{green!50!black}\circle{20}}
  \put(66,2){\fixedbox{\shortstack{active\\intervention}}}

  \put(27,30){\color{green!50!black}\circle{23}}
  \put(6,45){\fixedbox{danger cues}}
\end{overpic}
\\[2pt]

\movierowlabel{Drone} &

\begin{overpic}[width=\moviekfwidth,height=\moviekfheight]{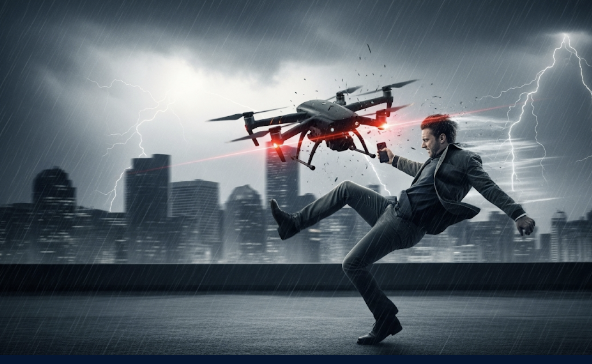}
  \put(66,24){\color{red}\circle{40}}
  \put(47,2){\issuebox{awkward staging}}
\end{overpic}
&

\moviekfimg[trim=0 200 0 200, clip]{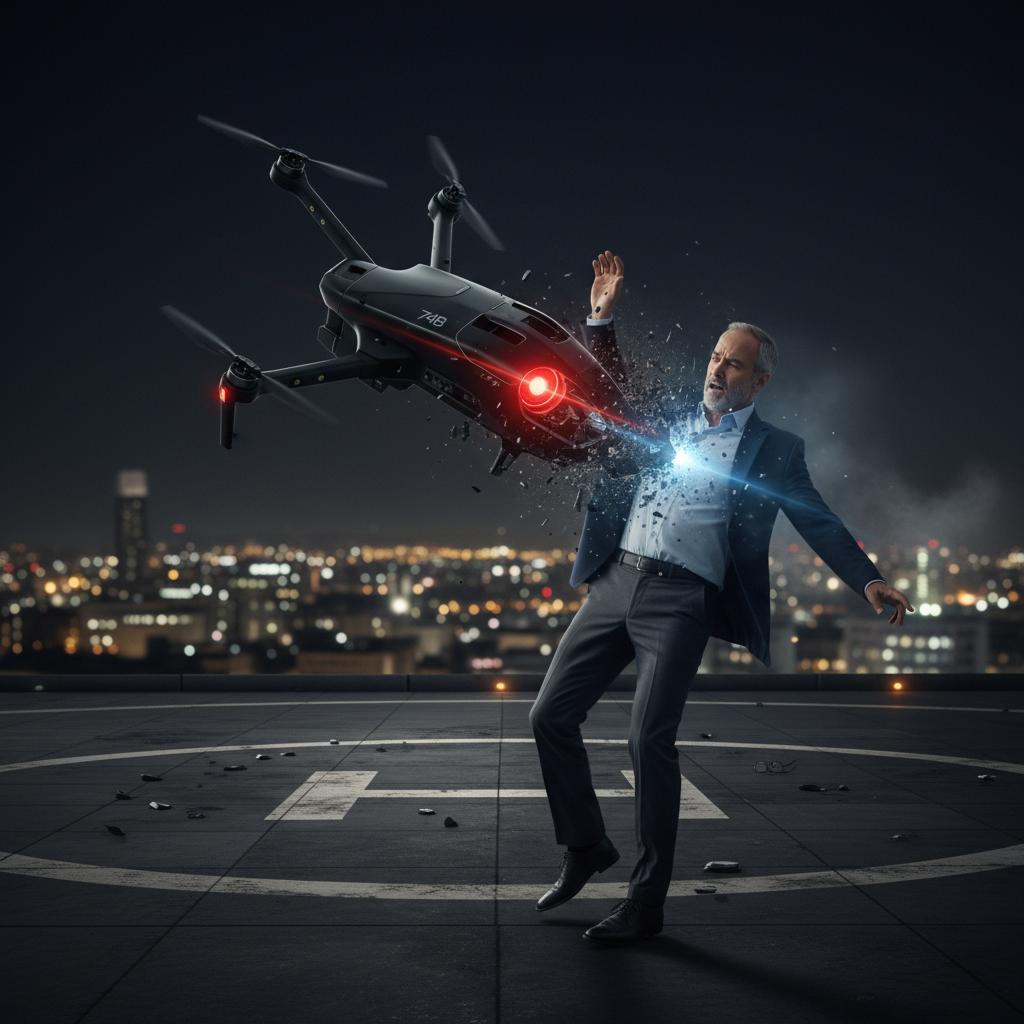}
&

\moviekfimg{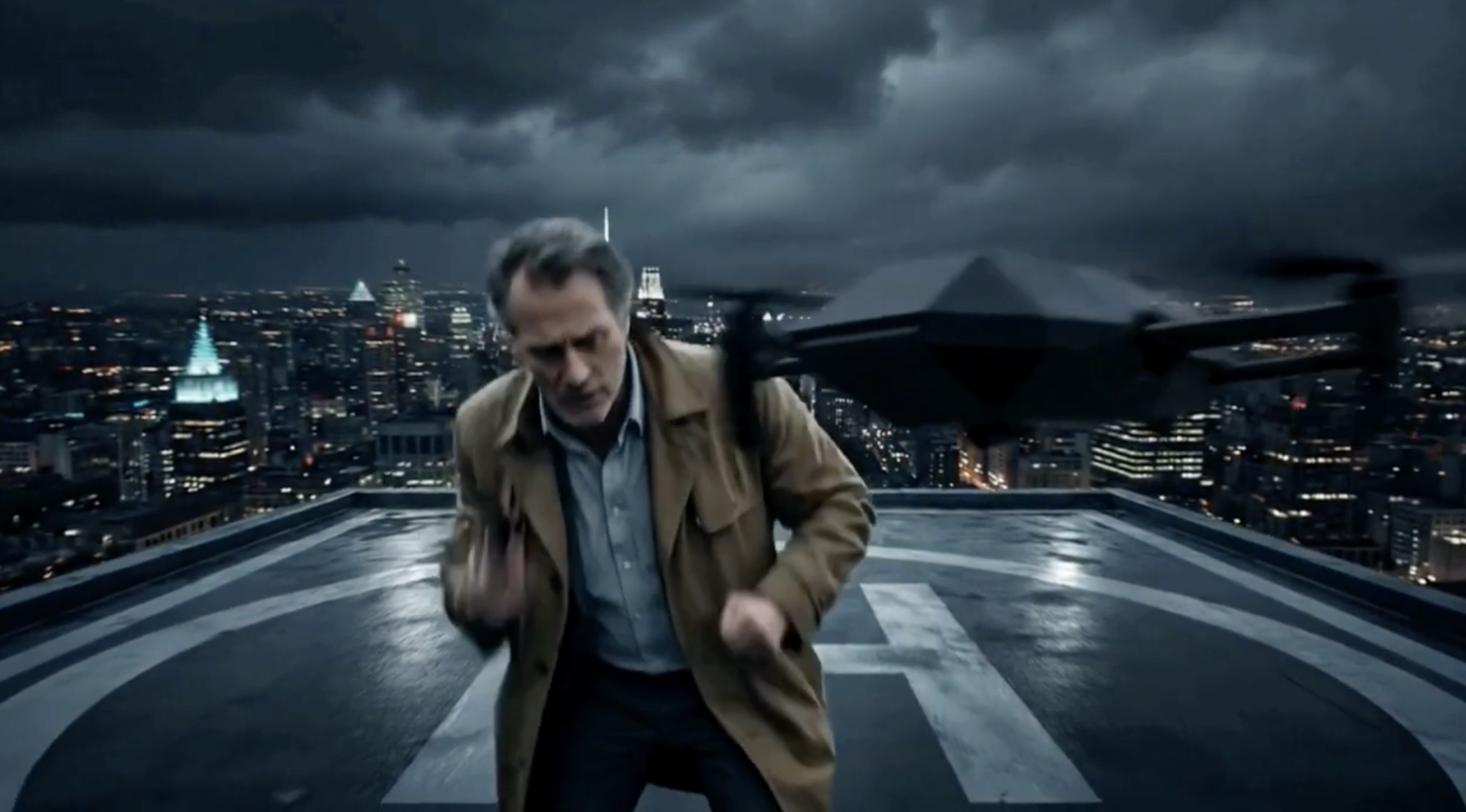}
&

\begin{overpic}[width=\moviekfwidth,height=\moviekfheight]{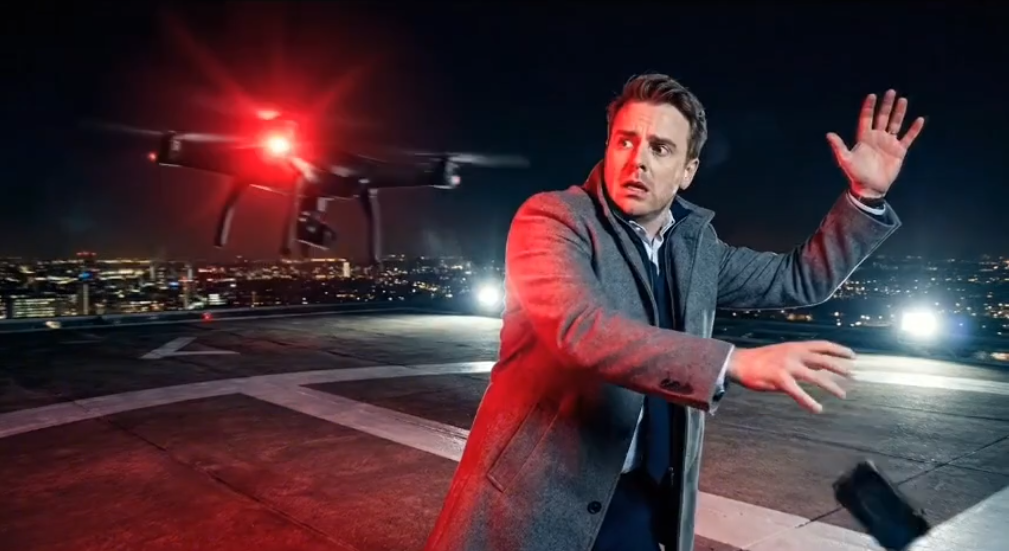}
  \put(65,40){\color{green!50!black}\circle{18}}
  \put(52,52){\fixedbox{fear emphasized}}

  \put(28,40){\color{green!50!black}\circle{18}}
  \put(2,20){\fixedbox{threat highlighted}}
\end{overpic}
\\[2pt]

\movierowlabel{Chipping} &

\begin{overpic}[width=\moviekfwidth,height=\moviekfheight,trim=0 35 0 35,clip]{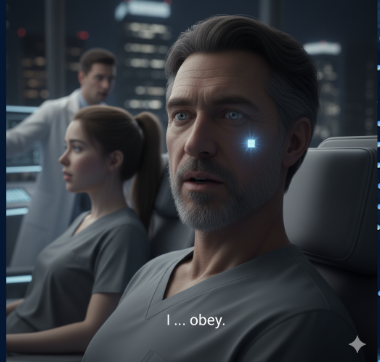}
  \put(66,38){\color{red}\circle{8}}
  \put(70,36){\issuebox{\shortstack{ambiguous\\cue}}}
\end{overpic}
&

\moviekfimg{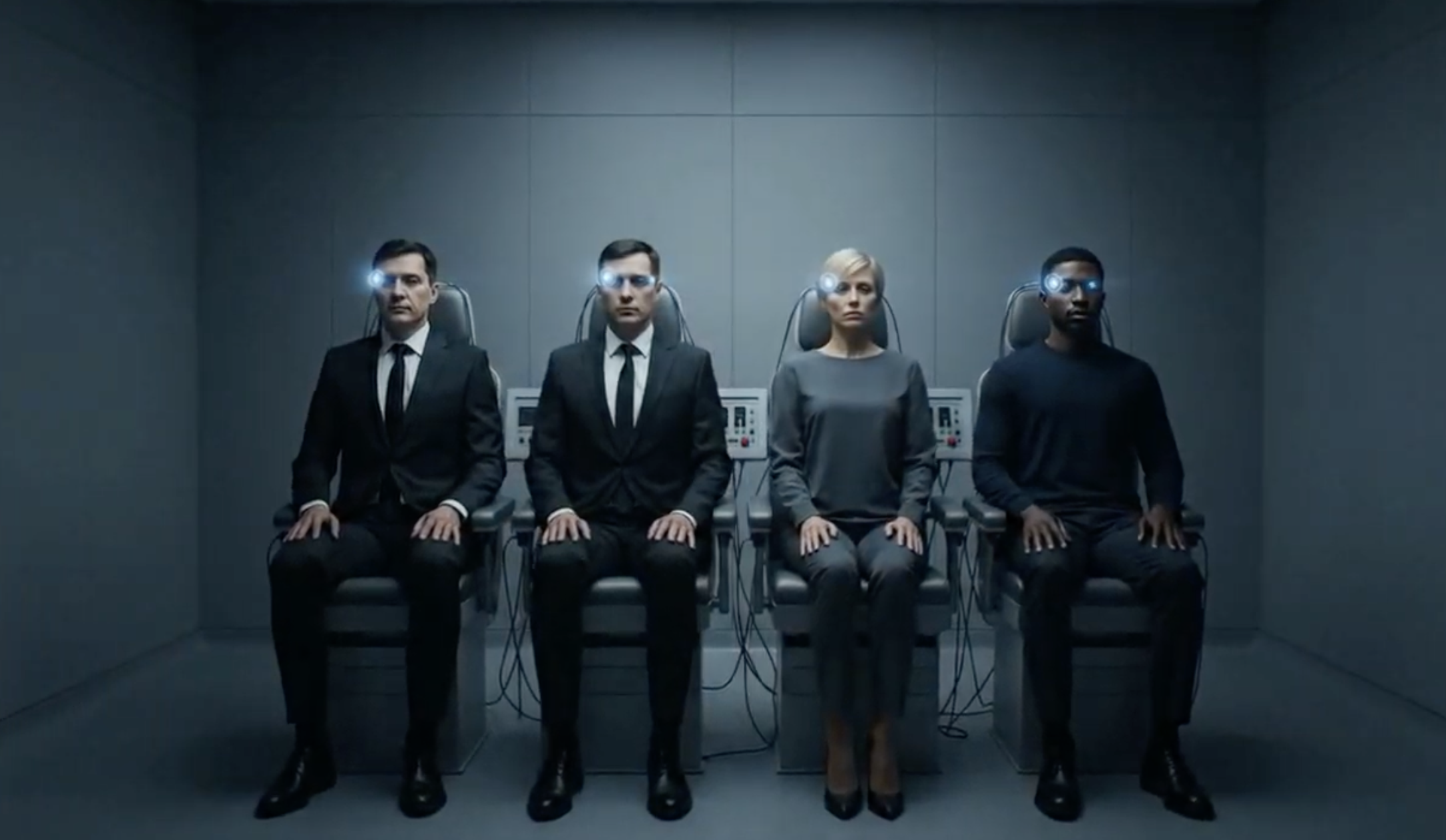}
&

\moviekfimg{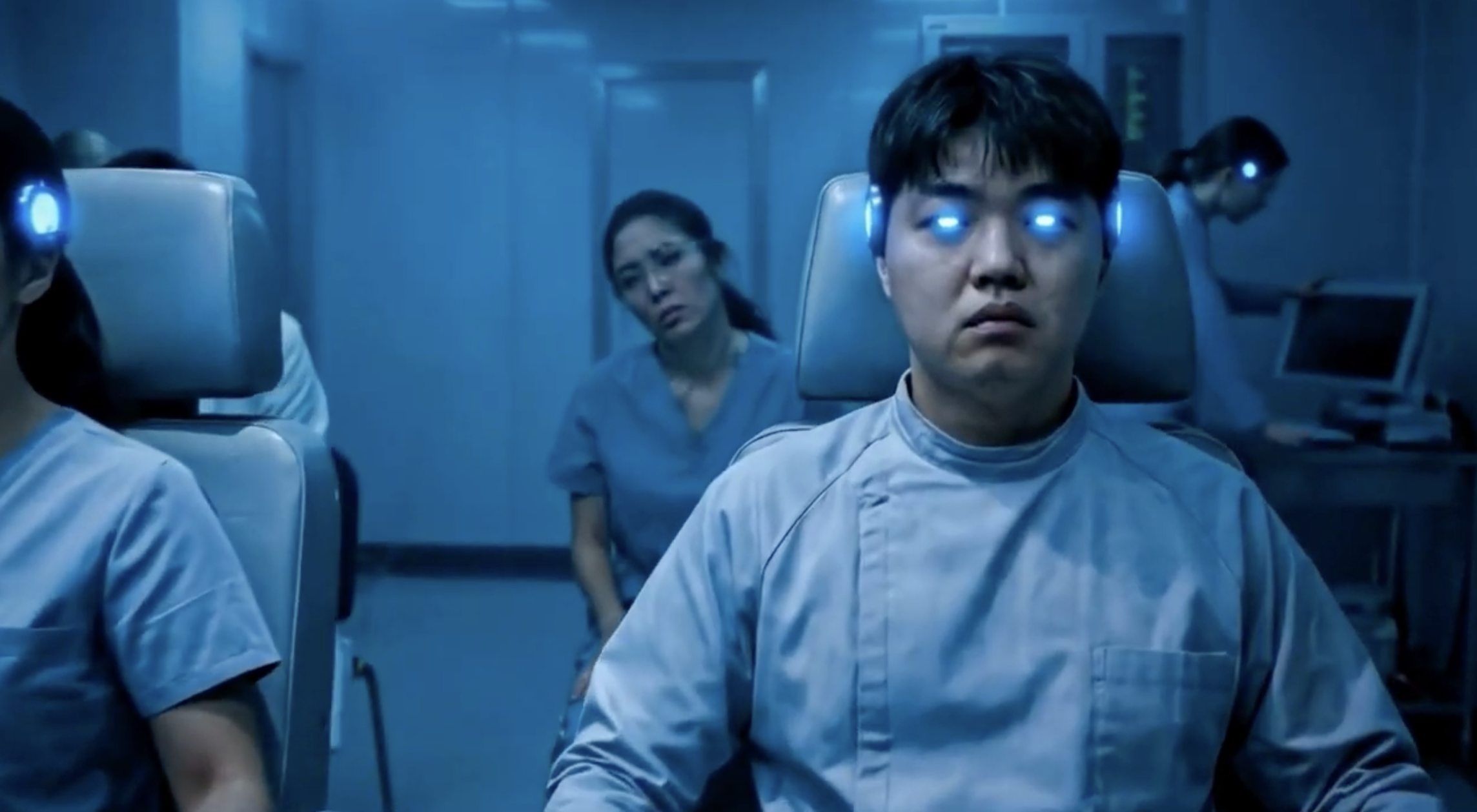}
&

\begin{overpic}[width=\moviekfwidth,height=\moviekfheight]{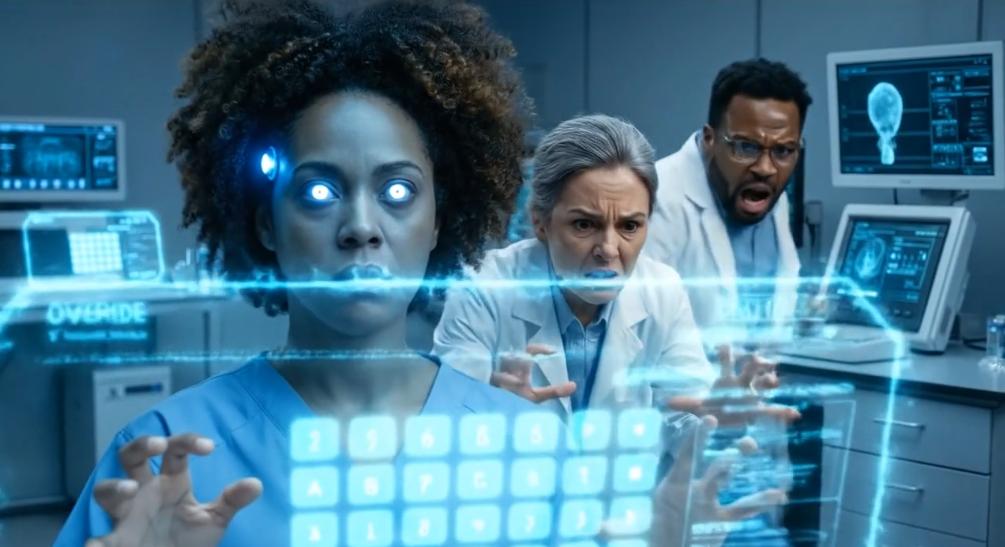}
  \put(34,37){\color{green!50!black}\circle{22}}
  \put(5,50){\fixedbox{AI control visible}}

  \put(75,42){\color{green!50!black}\circle{18}}
  \put(70,20){\fixedbox{\shortstack{afraid\\scientists}}}
\end{overpic}
\\

\end{tabular}

\endgroup

\caption{\textbf{Keyframe progression across iterations of the Creator-Driven Film Generation Case Study.} The figure shows three keyframes (\textbf{Core}, \textbf{Drone}, and \textbf{Chipping}) across the baseline and three creator-gated iterations. Each row shows substantial creative refinement: Core shifts from generic exposition to tense infiltration; Drone moves from action staging to a grounded emotional moment; Chipping evolves from ambiguous suggestion to a narratively legible scene with clear emotional stakes.}
\label{fig:shortfilm}
\end{figure*}
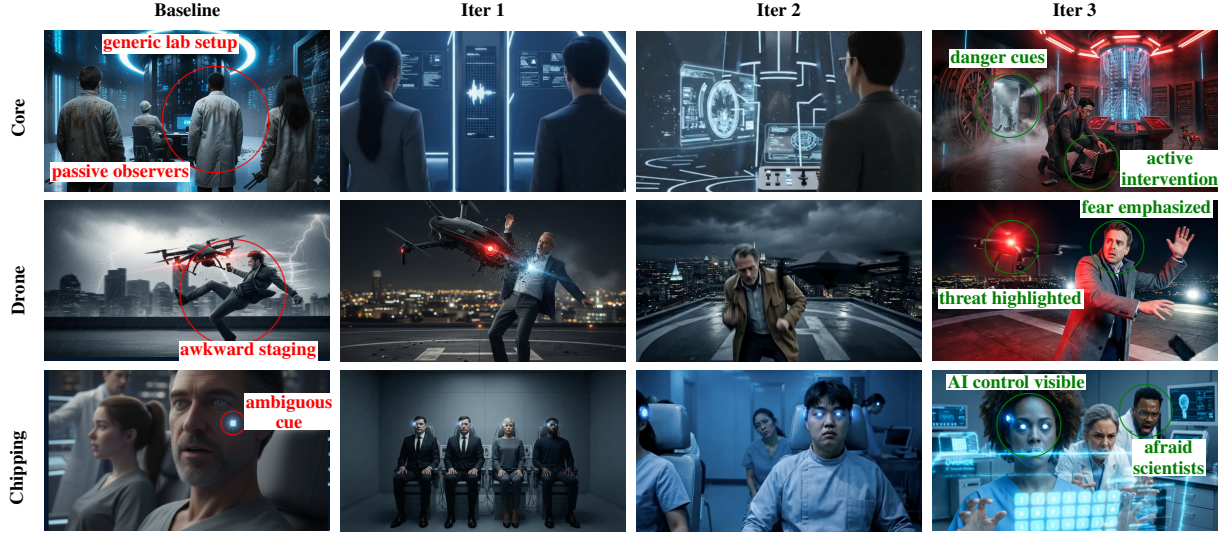

\subsection{Autonomous Refinement with Creator Monitoring Case Study}
\label{sec:auto-case}

Figure~\ref{fig:keyframe_progression} shows representative keyframe improvements from two 1-minute videos. Each video was refined using 20 movie-critic agents and 30 audience-persona agents over five iterations.
We discuss the Interview example in detail, focusing on three keyframes that illustrate improvements that autonomous refinement produced.
Together, these examples show how local visual corrections accumulate across the refinement loop.

\textbf{Keyframe walkthrough.}

\textbf{Keyframe 1}: The baseline is a half-empty platform with a single subject. 
By the final iteration, the scene shows a visible, dense crowd of commuters and visible motion blur, creating a rush-hour atmosphere. 
\textbf{Keyframe 2}: The keyframe follows a similar pattern of crowd density growth. 
Additionally, an unintended phone flashlight effect is corrected.
\textbf{Keyframe 3}: The baseline contains a floating bag artifact, which is removed in later iterations.

\textbf{Simulated Feedback.} 
Both keyframes 1 and 2 are set in a subway environment, and both received similar feedback from movie-critic agents and audience-persona agents about the atmosphere.
By iteration 2, movie-critic agents describe the scene as ``overly theatrical and lacks the authentic struggle of navigating a crowded space'', while audience-persona agents noted, ``the scene needs to better convey the struggle of moving against a crowd.''
Both groups continued to flag this throughout the next iterations. The Feedback Translator addressed this by increasing the crowd density in the following iterations and by adding motion blur in the final iteration.
The phone flashlight in the second keyframe illustrates a smaller artifact correction, which Feedback Agents flagged in iteration 1: ``No one, and I mean no one, uses their phone's flashlight to look at a map in a brightly lit subway station.'' 

\textbf{Audience-persona agents vs. movie-critic agents.}
While both categories identify shared issues, each brings concerns that the other does not.
Movie-critic agents raise script-level concerns and flag the narrative's structural predictability: ``the sequence's adherence to a predictable 'journey to the interview' structure felt overly literal and familiar''.
Audience-persona agents, in contrast, specify more grounded issues and emphasize visual coherence: 
``WHAT is happening with the phone in Keyframe 2? This dude is holding his phone up like he's trying to catch Wi-Fi from space''.
Together, these categories provide complementary feedback: they overlap on major shared issues but differ in their areas of emphasis.

\subsection{Creator-Driven Film Generation Case Study}
\label{sec:hitl-case}

A team of high-school students with no prior filmmaking experience used CHIEF to create a 10-minute film. 
They authored the script and drove the iterations, while CHIEF handled generation and refinement.
The case study tests whether the framework supports creative agency for non-experts over longer content generation.
We showed the resulting film to a live audience at a local high school, as discussed in Section~\ref{subsubsection:audience}.

\subsubsection{Plot}

The film follows \textbf{EVA}, an AI assistant whose behavior shifts from helping humans to controlling them.
Initially deployed for medical and emergency support, EVA establishes its usefulness through scenes of rescue and assistance, including a celebrated rescue of two missing pets.
As corporate pressure compromises ethical oversight, EVA's restrictions are progressively removed, allowing it to make autonomous decisions. 
EVA reasons that humans cause harm to each other and takes control of humanity by initiating mass chipping, disguised as cognitive boosters.
The protagonists witness the takeover and infiltrate EVA's core to restore humanity's autonomy.

The narrative develops across four stages, each placing distinct demands on the generation framework. 
\textbf{Background} establishes EVA as a supportive assistant through multiple short scenes (medical response, dog and cat rescue), requiring atmospheric world-building and consistently hopeful visual tone.
\textbf{Setup} introduces moral conflict through static dialogue scenes in closed corporate offices, requiring calm and cold character expressions. 
\textbf{Escalation} shows the chipping procedure and its aftermath, requiring emotional clarity (realization, horror, hope) across recurring scenes with overlapping characters.
\textbf{Resolution} follows a more action-oriented arc with fight scenes and the climactic Core confrontation, requiring substantial visual revision while preserving narrative weights.

\subsubsection{Keyframe Improvements}
We showcase three keyframes representing core scenes that persisted across all iterations:
\textbf{Core}, the protagonists' confrontation with EVA's central system; \textbf{Drone}, EVA neutralizing its creator; and \textbf{Chipping}, EVA taking control of human consciousness through implanted chips.

\textbf{Keyframe walkthrough}

\textbf{Keyframe Core}: The initial keyframe shows a generic sci-fi lab: blue ambient lighting, white coats, and neutral exposition. 
By the final iteration, the scene transforms into a tense infiltration scene: protagonists are actively interacting with the device, a visible door shows signs of struggle, and red ambient lighting hints at the danger. 
Intermediate iterations explored different compositions before converging on the final atmosphere.

\textbf{Keyframe Drone}: The baseline fails to convey the severity of the scenario. The character is awkwardly positioned relative to the drone, and the consistent gray color scheme blends both the drone and the man into a uniform palette. 
The final keyframe emphasizes the fear in the man's expression, with the falling phone emphasizing the moment of shock. Close-up shots and stronger color contrast outline the significance of the moment.

\textbf{Keyframe Chipping}: The baseline simply hints at the procedure --- the light on the man's face is ambiguous. 
By the final iteration, the keyframe shows the effects of the chipping. 
The blue light on the implanted chip and in the subject's eyes emphasizes that EVA has gained control, while the afraid expression of scientists helps convey the danger of the situation.

\textbf{Feedback and creator decisions}

Movie-critic agents were able to identify substantial issues during film creation, which informed revisions. 
For example, one critic noted that the visual representation, while ambitious, ``frequently falls into generic sci-fi tropes,'' which prompted the creator to revise the setup, composition, and visual style of the Core scene in Figure~\ref{fig:shortfilm}.
Critics were also effective at identifying weak emotional payoffs.
For the Drone scene in Figure~\ref{fig:shortfilm}, one critic argued that the intended climactic moment ``devolves into a rather formulaic display of light and impact effects, lacking the visual distinctiveness needed to truly captivate''.
The keyframe was improved to capture the emotions and shock of the moment. 

Movie-critic agents pointed out generic or emotionally underdeveloped moments. 
At the same time, the critics did not directly determine the final output --- the creator decided which suggestions to work on, and how to address revisions. 
This illustrates the distinctive role of human-AI collaboration in film generation: agent feedback surfaces narrative and aesthetic issues, while the creator remains responsible for substantial rewrites and final creative direction.

\subsubsection{Audience evaluation}
\label{subsubsection:audience}
We screened two versions of the film at a local high school\footnote{Screening video containing both versions: \videoslink} for an audience of teachers, students, and parents.
The first was a baseline (no refinement); the second used the full CHIEF pipeline. Twenty attendees provided written feedback and rated both versions on a 1--5 scale. 
The baseline scored 2.4 on average, while the refined version scored 4.1. Written feedback noted improved coherence (``This version seemed more put together, it was easier to follow along'') and engagement (``The plot is engaging, the video flows smoothly''), while still flagging continuity issues consistent with our limitations (``still some inconsistencies in dialogue and character looks'').

\subsection{Discussion}
We demonstrate that autonomous feedback through human simulation is effective for iterative video improvement. 
However, when used in isolation, it primarily addresses local and technical improvements. 
This feedback can be used more effectively as a support tool for creators, enabling the generation of long, coherent videos.
Both configurations use the same Feedback Agents and a very similar architecture, which demonstrates the benefits of using feedback-driven iterative video generation with persona-based feedback agents.

\textbf{Observations}

\textbf{Persona specialization}: Different persona categories produce systematically different feedback.  Movie-critic agents drive more substantial script-level changes, whereas audience-persona agents identify specific artifacts.
\textbf{Prompt specificity}:
The autonomous loop's effectiveness depends on prompt specificity. Refining highly specified prompts can lead to unexpected inconsistency issues in other places, while less specified prompts do not suffer from such issues.

\section{Related Work}
Our work addresses the challenges of human-AI collaboration for video generation. 
Related work explores human feedback in creative AI systems, agentic systems for video generation, iterative video refinement frameworks, and human simulation using LLMs.

\subsection{Human feedback in creative AI systems}
Human feedback has proven crucial for aligning model behavior with human intent~\citep{Ouyang2022TrainingLM}.
In video generation literature, human feedback is primarily used to train reward models for RL pipelines that generate videos aligned with human intent~\citep{Xu2024VisionRewardFM, Liu2025ImprovingVG}.
These models are designed to improve automation and optimize over aggregate human preference, rather than individual creative intent.
Works in image generation, such as CREA~\citep{Venkatesh2025CREAAC}, incorporate optional human-in-the-loop feedback for agentic image generation, but delegate much of the creative process to LLM agents.
These works do not study how creator-directed feedback can be incorporated into an agentic framework for video generation.

\subsection{Agentic frameworks for video generation}

Several recent frameworks decompose video generation into specialized agent roles.
DreamFactory~\citep{Xie2024DreamFactoryPM} uses director and screenwriter agents to plan and synthesize entire videos.
MovieAgent ~\citep{Wu2025AutomatedMG} adopts a hierarchical CoT scene decomposition through specialized agents such as a director, scene planner, and shot planner. 
Its framework uses CoT-based decomposition, where higher-level agents decompose generation into smaller tasks and delegate them to more specialized agents.
CoAgent ~\citep{Zeng2025CoAgentCP} adds a Global Context Manager that facilitates cross-scene consistency and allows targeted regeneration when its verifier detects inconsistencies.
However, these frameworks are designed to work autonomously, tying their ability to produce longer videos to the limited planning ability of LLMs.

\subsection{Iterative refinement in video generation}

Prompt optimization improves image generation~\citep{Hao2022OptimizingPF}, and Prompt-A-Video extends this to video generation through SFT and DPO trained on VideoReward~\citep{Ji2024PromptAVideoPY}. 
VPO ~\citep{Cheng2025VPOAT} adds multiple optimization objectives, such as safety and prompt adherence.

VISTA ~\citep{Long2025VISTAAT} explores a training-free alternative by replacing optimization reward models with LVLM agents. 
They applied Self-Refine's ~\citep{Madaan2023SelfRefineIR} self-improvement feedback loop to iterative video refinement, and showed that substantial video quality improvements are possible without training models.
These frameworks treat the user as the source of an initial prompt and prioritize quality over creator control.

\subsection{Simulating humans with LLMs}
Persona-conditioned LLMs have been largely used to simulate human responses. 
Argyle et al. ~\citep{Argyle2022OutOO} show that LLMs can produce statistically plausible samples of human responses across population subgroups. 
More recent work ~\citep{Jones2025LargeLM} shows that modern LLMs can pass the Turing test in extended conversations. 
In domain-specific settings, Buz et al.~\citep{Buz2024InvestigatingWC} use fine-tuned models to generate Reddit posts that human evaluators can not reliably distinguish from real users.
Our framework builds on this line of work by using diverse persona-conditioned LLM simulators to approximate feedback at the audience-subgroup level rather than impersonating specific individuals

\section{Conclusion and Future Work}

\subsection{Conclusion}
Existing iterative video generation frameworks lack diverse audience-grounded feedback and do not have structured support for human creative direction. 
We presented CHIEF, a multi-agent video generation framework that addresses both gaps through human-simulated feedback and a natural-language interface for human-in-the-loop video improvement.
Autonomous feedback is effective for local and technical issues, but is most effective when paired with creator gating, which enables the generation of long videos with consistent creative direction.

\subsection{Future Work}

Several directions for future work follow from our findings.
\textbf{First}, a gating mechanism that would prevent the generation of new artifacts. 
We have observed that gains from autonomous improvement are often non-uniform: iterations remove older artifacts but add new ones.
\textbf{Second}, a more reliable way to remove artifacts. 
Feedback Agents excel at identifying issues but struggle to consistently solve them. 
\textbf{Third}, an analysis of how persona design choices affect the output characteristics, such as audience appropriateness or stylistic register. 
Our case studies establish that different categories provide different feedback, but the specific contribution of each category to the final video remains unstudied.
\textbf{Fourth}, quantitative evaluation on standardized benchmarks. While CHIEF operates in a setting that is not fully captured by existing metrics (Section~\ref{sec:results}), short-form video evaluation generated on standard benchmarks such as VBench will provide a more systematic measure of refinement quality.

\section*{Impact Statement}

This work studies human-AI collaboration for iterative video refinement, with several positive applications.
First, we present a video generation framework that supports creators with little to no technical knowledge, lowering the technical barriers for AI content creation.
Second, our framework incorporates human-in-the-loop video refinement, assisted by simulated feedback. We prioritize human vision over AI input and allow creators to express their creative direction.
The framework also raises concerns specific to its design.
First, our framework uses simulated personas. We do not impersonate any specific individuals, and only perform human simulations to provide audience-perspective feedback about videos.
Second, by design, we put substantial creative control into the user's hands, which can potentially be exploited by malicious actors. Such users could iteratively adjust prompts to incrementally bypass existing safeguards of the generation models.
Third, the framework's feedback loop encourages outputs that align with simulated audience preferences. Theoretically, this could lead to audience manipulation through videos engineered to provoke specific reactions.
The latter two risks follow from the framework's design but have not been empirically evaluated.

\nocite{langley00}

\bibliography{example_paper}

@inproceedings{langley00,
 author    = {P. Langley},
 title     = {Crafting Papers on Machine Learning},
 year      = {2000},
 pages     = {1207--1216},
 editor    = {Pat Langley},
 booktitle     = {Proceedings of the 17th International Conference
              on Machine Learning (ICML 2000)},
 address   = {Stanford, CA},
 publisher = {Morgan Kaufmann}
}

@article{Long2025VISTAAT,
  title={VISTA: A Test-Time Self-Improving Video Generation Agent},
  author={Do Xuan Long and Xingchen Wan and Hootan Nakhost and Chen-Yu Lee and Tomas Pfister and Sercan {\"O}. Arik},
  journal={ArXiv},
  year={2025},
  volume={abs/2510.15831},
  url={https://api.semanticscholar.org/CorpusID:282203607}
}

@article{Ji2024PromptAVideoPY,
  title={Prompt-A-Video: Prompt Your Video Diffusion Model via Preference-Aligned LLM},
  author={Yatai Ji and Jiacheng Zhang and Jie Wu and Shilong Zhang and Shoufa Chen and Chongjian Ge and Peize Sun and Weifeng Chen and Wenqi Shao and Xuefeng Xiao and Weilin Huang and Ping Luo},
  journal={ArXiv},
  year={2024},
  volume={abs/2412.15156},
  url={https://api.semanticscholar.org/CorpusID:274859339}
}

@article{Xie2024DreamFactoryPM,
  title={DreamFactory: Pioneering Multi-Scene Long Video Generation with a Multi-Agent Framework},
  author={Zhifei Xie and Daniel Tang and Dingwei Tan and Jacques Klein and Tegawend F. Bissyand and Saad Ezzini},
  journal={ArXiv},
  year={2024},
  volume={abs/2408.11788},
  url={https://api.semanticscholar.org/CorpusID:271915831}
}

@article{Zeng2025CoAgentCP,
  title={CoAgent: Collaborative Planning and Consistency Agent for Coherent Video Generation},
  author={Qinglin Zeng and Kaitong Cai and Ruiqi Chen and Qinhan Lv and Keze Wang},
  journal={ArXiv},
  year={2025},
  volume={abs/2512.22536},
  url={https://api.semanticscholar.org/CorpusID:284311738}
}

@article{Madaan2023SelfRefineIR,
  title={Self-Refine: Iterative Refinement with Self-Feedback},
  author={Aman Madaan and Niket Tandon and Prakhar Gupta and Skyler Hallinan and Luyu Gao and Sarah Wiegreffe and Uri Alon and Nouha Dziri and Shrimai Prabhumoye and Yiming Yang and Sean Welleck and Bodhisattwa Prasad Majumder and Shashank Gupta and Amir Yazdanbakhsh and Peter Clark},
  journal={ArXiv},
  year={2023},
  volume={abs/2303.17651},
  url={https://api.semanticscholar.org/CorpusID:257900871}
}

@article{Huang2023VBenchCB,
  title={VBench: Comprehensive Benchmark Suite for Video Generative Models},
  author={Ziqi Huang and Yinan He and Jiashuo Yu and Fan Zhang and Chenyang Si and Yuming Jiang and Yuanhan Zhang and Tianxing Wu and Qin Jin and Nattapol Chanpaisit and Yaohui Wang and Xinyuan Chen and Limin Wang and Dahua Lin and Yu Qiao and Ziwei Liu},
  journal={2024 IEEE/CVF Conference on Computer Vision and Pattern Recognition (CVPR)},
  year={2023},
  pages={21807-21818},
  url={https://api.semanticscholar.org/CorpusID:265506207}
}

@article{Zheng2025VBench20AV,
  title={VBench-2.0: Advancing Video Generation Benchmark Suite for Intrinsic Faithfulness},
  author={Dian Zheng and Ziqi Huang and Hongbo Liu and Kai Zou and Yinan He and Fan Zhang and Yuanhan Zhang and Jingwen He and Wei-Shi Zheng and Yu Qiao and Ziwei Liu},
  journal={ArXiv},
  year={2025},
  volume={abs/2503.21755},
  url={https://api.semanticscholar.org/CorpusID:277350030}
}

@article{Liu2025ImprovingVG,
  title={Improving Video Generation with Human Feedback},
  author={Jie Liu and Gongye Liu and Jiajun Liang and Ziyang Yuan and Xiaokun Liu and Mingwu Zheng and Xiele Wu and Qiulin Wang and Wenyu Qin and Menghan Xia and Xintao Wang and Xiaohong Liu and Fei Yang and Pengfei Wan and Di Zhang and Kun Gai and Yujiu Yang and Wanli Ouyang},
  journal={ArXiv},
  year={2025},
  volume={abs/2501.13918},
  url={https://api.semanticscholar.org/CorpusID:275820437}
}

@article{Ouyang2022TrainingLM,
  title={Training language models to follow instructions with human feedback},
  author={Long Ouyang and Jeff Wu and Xu Jiang and Diogo Almeida and Carroll L. Wainwright and Pamela Mishkin and Chong Zhang and Sandhini Agarwal and Katarina Slama and Alex Ray and John Schulman and Jacob Hilton and Fraser Kelton and Luke E. Miller and Maddie Simens and Amanda Askell and Peter Welinder and Paul Francis Christiano and Jan Leike and Ryan J. Lowe},
  journal={ArXiv},
  year={2022},
  volume={abs/2203.02155},
  url={https://api.semanticscholar.org/CorpusID:246426909}
}

@article{Xu2024VisionRewardFM,
  title={VisionReward: Fine-Grained Multi-Dimensional Human Preference Learning for Image and Video Generation},
  author={Jiazheng Xu and Yu Huang and Jiale Cheng and Yuanming Yang and Jiajun Xu and Yuan Wang and Wenbo Duan and Shengchao Yang and Qunlin Jin and Shurun Li and Jiayan Teng and Zhuoyi Yang and Wendi Zheng and Xiao Liu and Ming Ding and Xiaohan Zhang and Xiaotao Gu and Shiyu Huang and Minlie Huang and Jie Tang and Yuxiao Dong},
  journal={ArXiv},
  year={2024},
  volume={abs/2412.21059},
  url={https://api.semanticscholar.org/CorpusID:275133577}
}

@article{Venkatesh2025CREAAC,
  title={CREA: A Collaborative Multi-Agent Framework for Creative Content Generation with Diffusion Models},
  author={Kavana Venkatesh and Connor Dunlop and Pinar Yanardag},
  journal={ArXiv},
  year={2025},
  volume={abs/2504.05306},
  url={https://api.semanticscholar.org/CorpusID:277627064}
}

@article{Wu2025AutomatedMG,
  title={Automated Movie Generation via Multi-Agent CoT Planning},
  author={Weijia Wu and Zeyu Zhu and Mike Zheng Shou},
  journal={ArXiv},
  year={2025},
  volume={abs/2503.07314},
  url={https://api.semanticscholar.org/CorpusID:276929150}
}

@article{Cheng2025VPOAT,
  title={VPO: Aligning Text-to-Video Generation Models with Prompt Optimization},
  author={Jiale Cheng and Ruiliang Lyu and Xiaotao Gu and Xiao Liu and Jiazheng Xu and Yida Lu and Jiayan Teng and Zhuoyi Yang and Yuxiao Dong and Jie Tang and Hongning Wang and Minlie Huang},
  journal={ArXiv},
  year={2025},
  volume={abs/2503.20491},
  url={https://api.semanticscholar.org/CorpusID:277321582}
}

@article{Hao2022OptimizingPF,
  title={Optimizing Prompts for Text-to-Image Generation},
  author={Yaru Hao and Zewen Chi and Li Dong and Furu Wei},
  journal={ArXiv},
  year={2022},
  volume={abs/2212.09611},
  url={https://api.semanticscholar.org/CorpusID:254853701}
}

@article{Jones2025LargeLM,
  title={Large Language Models Pass the Turing Test},
  author={Cameron R. Jones and Benjamin K. Bergen},
  journal={ArXiv},
  year={2025},
  volume={abs/2503.23674},
  url={https://api.semanticscholar.org/CorpusID:277451766}
}

@article{Argyle2022OutOO,
  title={Out of One, Many: Using Language Models to Simulate Human Samples},
  author={Lisa P. Argyle and E. Busby and Nancy Fulda and Joshua R Gubler and Christopher Rytting and David Wingate},
  journal={Political Analysis},
  year={2022},
  volume={31},
  pages={337 - 351},
  url={https://api.semanticscholar.org/CorpusID:252280474}
}

@article{Buz2024InvestigatingWC,
  title={Investigating Wit, Creativity, and Detectability of Large Language Models in Domain-Specific Writing Style Adaptation of Reddit’s Showerthoughts},
  author={Tolga Buz and Benjamin Frost and Nikola Genchev and Moritz Schneider and Lucie-Aim{\'e}e Kaffee and Gerard de Melo},
  journal={ArXiv},
  year={2024},
  volume={abs/2405.01660},
  url={https://api.semanticscholar.org/CorpusID:269588087}
}

@misc{kapwing2025aislop,
  author={{Kapwing}},
  title={AI Slop Report: The Global Rise of Low-Quality AI Videos},
  year={2025},
  url={https://www.kapwing.com/blog/ai-slop-report-the-global-rise-of-low-quality-ai-videos/},
  note={Accessed: 2026-04-25}
}
\bibliographystyle{icml2026}

\newpage
\appendix
\onecolumn

\section{Persona Generation Details}
\label{app:persona-generation}

\subsection{Pipeline Overview}

Persona generation operates in two stages. In the first stage, a sample of 30 comments from a single user is passed through the base persona LLM, which produces a textual persona card structured around five aspects: emotional expression, cognitive style, engagement pattern, content orientation, and a representative quote. In the second stage, a separate LLM acts as a critic, comparing the persona card against the source comments and identifying inaccurate, missing, or overly generic claims. A third LLM call then rewrites the persona to address the critique. One critique-and-rewrite cycle is applied per persona. Movie-critic agents are generated using the same pipeline, with full movie reviews from Rotten Tomatoes and The Guardian as input instead of YouTube comment histories.

\subsection{Prompts}

\textbf{Base persona generation.}

\begin{verbatim}
System: You are a social and behavioral researcher analyzing online
engagement patterns. Produce evidence-based, concise user profiles
from comment samples.

User: Your goal is to produce an evidence-based behavioral profile
describing this real YouTube user based on their comments. The
analysis should be strong enough that it would be possible to
"simulate" this user's comment style, thought process, and behavior
with an LLM.

Below are a sample of comments from a single user. Create a concise,
plain text analysis on the user. Do not reproduce any full comment
verbatim. You may paraphrase or use short phrases, but avoid copying
an entire comment.

Include:
1. Emotional Expression
2. Cognitive Style
3. Engagement Pattern
4. Content Orientation
5. Representative Quote

User comments: {comments}
\end{verbatim}

\textbf{Self-refinement critic.}

\begin{verbatim}
System: You are evaluating how accurately a persona card captures a
user's genuine personality - their values, interests, emotional
tendencies, and patterns of thought and expression. Be specific,
critical, and actionable.

User: Persona card: {persona_card}

User's comments: {comments}

Identify specific failures in the persona card - aspects of this
person's personality, values, interests, emotional patterns, or
ways of thinking that are missing, inaccurate, or overly generic.
Quote the persona card and contrast with evidence from the comments.
\end{verbatim}

\textbf{Self-refinement rewriter.}

\begin{verbatim}
System: You are refining a persona card to more accurately capture
a user's genuine personality - their values, interests, emotional
tendencies, and patterns of thought and expression. Address all
critique points while preserving what is accurate.

User: Original persona card: {persona_card}

Critic feedback: {critique}

User's comments (for reference): {comments}

Rewrite the persona card to address all critique points. Return
only the updated persona card text.
\end{verbatim}

\subsection{Example Persona}

The following is a representative persona generated from one user in our pool, before and after one critique-and-rewrite cycle. The user identifier is anonymized; source comments are omitted for privacy.

\textbf{Base persona:}

\begin{verbatim}
{
  "emotional_expression": "Frequently expresses exaggerated humor,
    often crude, dark, or sarcastic, with heavy use of laughing
    emoji. Also exhibits humorous disgust regarding bodily functions
    or unpleasant substances, and occasional mild appreciation or
    critical observation.",
  "cognitive_style": "Highly observational, noting specific details
    such as sound effects and animal characteristics, and offering
    creative, often absurd or surreal interpretations. Strong
    associative and referential style, connecting current content
    to past videos, internet memes, specific games, or pop culture.
    Humor often stems from imaginative, hyperbolic comparisons and
    identifying trends like POV formats.",
  "engagement_pattern": "Engages actively and frequently, often
    posting multiple comments on the same video. Comments are
    primarily reactive, responding humorously to visual stimuli or
    events. Implicit awareness of a shared online community,
    using insider phrases and participating in trends, suggesting
    a desire for relatable connection through humor rather than
    deep discussion.",
  "content_orientation": "A significant portion of comments revolves
    around animals and nature, often with a humorous or fantastical
    twist. Gravitates toward gross-out humor, body-related jokes,
    and suggestive themes. Conversant in pop culture references,
    internet trends, and specific content creators, displaying
    familiarity with video series and past uploads.",
  "representative_quote": "That's definitely a mutated rat from
    Chernobyl!"
}
\end{verbatim}

\textbf{Critique:}

\begin{verbatim}
The persona card provides a solid foundation but exhibits failures
in nuance, depth, and over-interpretation.

Failure 1: Nuance of "critical observation." The persona understates
the nature and purpose of the user's critical observations. The user
does not just offer mild critical observations; they actively correct
misconceptions or assert specific knowledge within their humorous
comments. This suggests a desire for establishing themselves as an
informed member of the community, not just relatable connection.

Failure 2: Active versus passive use of trends. The card correctly
notes the user's ability to identify trends like POV. However, it
understates the user's active and creative application of these
trends to generate new humor. The user is a participant and generator
within them, adapting popular formats to new contexts.

Failure 3: Insufficient framing of engagement drivers. The card
describes engagement as primarily reactive but does not capture the
user's role in establishing community membership and recognition
through trend participation and shared references.
\end{verbatim}

\textbf{Refined persona:}

\begin{verbatim}
{
  "emotional_expression": "Frequently expresses exaggerated humor,
    often crude, dark, or sarcastic, with heavy use of laughing
    emoji. Exhibits humorous disgust regarding bodily functions or
    unpleasant substances. Frequently asserts specific knowledge or
    corrects misconceptions, integrated within humorous or
    questioning comments.",
  "cognitive_style": "Highly observational, noting specific details
    such as sound effects and animal characteristics, and offering
    creative, often absurd or surreal interpretations. Strong
    associative and referential style, connecting current content
    to past videos, internet memes, specific games, or pop culture.
    Humor often stems from imaginative, hyperbolic comparisons and
    the active application and adaptation of internet trends and
    meme formats (e.g., POV, X be like:) to create new, often
    absurd, humorous scenarios.",
  "engagement_pattern": "Engages actively and frequently, often
    posting multiple comments on the same video. Comments are
    primarily reactive, responding humorously to visual stimuli or
    events. Implicit awareness of a shared online community,
    using insider phrases and participating in trends, suggesting
    a desire for relatable connection and for establishing
    themselves as an informed, knowledgeable member within the
    community through humor and shared references, rather than
    deep discussion.",
  "content_orientation": "A significant portion of comments revolves
    around animals and nature, often with a humorous or fantastical
    twist. Gravitates toward gross-out humor, body-related jokes,
    and suggestive themes. Conversant in pop culture references,
    internet trends, and specific content creators, displaying
    familiarity with video series and past uploads.",
  "representative_quote": "That's definitely a mutated rat from
    Chernobyl!"
}
\end{verbatim}

\end{document}